\documentclass{egpubl}
\usepackage{pg2024}

\SpecialIssueSubmission    

\usepackage[T1]{fontenc}
\usepackage{dfadobe}

\usepackage{cite}  
\BibtexOrBiblatex
\electronicVersion
\PrintedOrElectronic
\ifpdf \usepackage[pdftex]{graphicx} \pdfcompresslevel=9
\else \usepackage[dvips]{graphicx} \fi
\usepackage{bbding}
\usepackage{amsmath}
\usepackage{egweblnk} 
\usepackage{amssymb} 
\usepackage{booktabs} 

\usepackage{orcidlink}
\usepackage{soul}
\usepackage{makecell}
\usepackage{colortbl}
\usepackage{xcolor}
\usepackage{multirow}

\usepackage{amssymb}
\usepackage{multirow}
\usepackage{bbding}
\usepackage{color}

\title[SCARF]{SCARF: Scalable Continual Learning Framework for \\ Memory-efficient Multiple Neural Radiance Fields}

\author[Y. Wang et al.]{\parbox{\textwidth}{\centering Yuze Wang $^{1}$\orcid{0009-0000-7676-3408}, Junyi Wang$^{2,1}$\orcid{0000-0002-3191-1662}, Chen Wang$^{3}$\orcid{0000-0002-4334-6103}, Wantong Duan$^{4,1}\orcid{0009-0001-8238-0670}$, Yongtang Bao$^{5}\orcid{0000-0002-1010-7229}$ and Yue Qi\footnotemark[1]$^{1}\orcid{0000-0001-9304-1933}$
        }
        \\
{\parbox{\textwidth}{\centering $^1$State Key Laboratory of Virtual Reality Technology and Systems, School of Computer Science and Engineering, Beihang University, $^2$School of Computer Science and Technology, Shandong University, $^3$School of Computer and Artificial Intelligence, Beijing Technology and Business University,$^4$Jingdezhen Research Institute of Beihang University, $^5$Shandong University of Science and Technology, \footnotemark[1] Corresponding Author
       }
}
}

\begin{document}

\teaser{
 \includegraphics[width=1.0\linewidth]{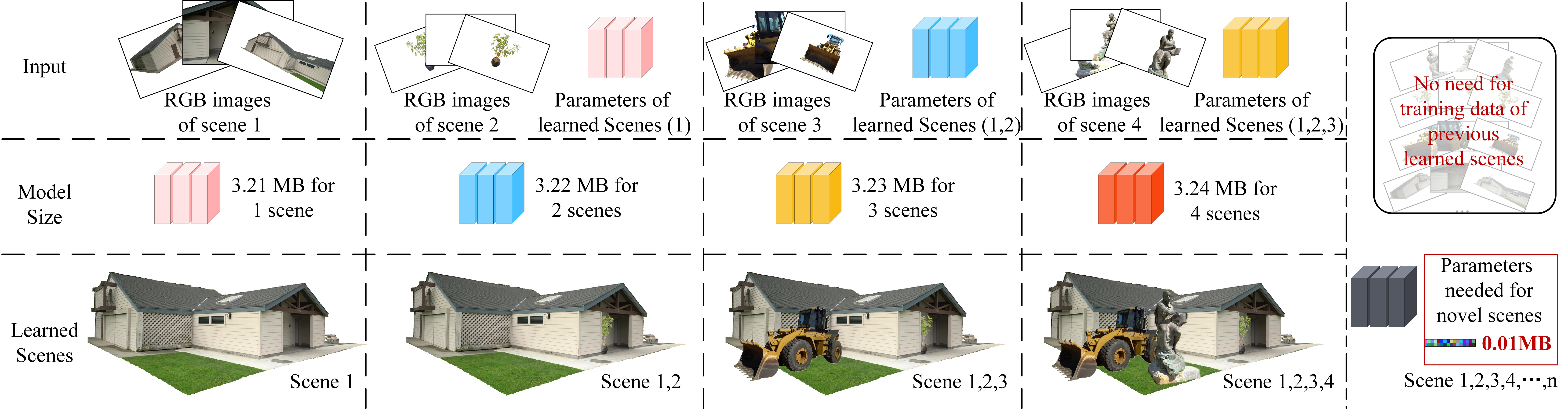}
 \centering
  \caption{Our method is able to incrementally learn multiple 3D scenes with only the training data of the upcoming new scene available, no need for training data of previous learned scenes, and represent them in a memory-efficient way. Only 0.01MB of additional storage is required to train each new 3D scene.}
\label{teaser}
}

\maketitle
\begin{abstract}
    This paper introduces a novel continual learning framework for synthesising novel views of multiple scenes, learning multiple 3D scenes incrementally, and updating the network parameters only with the training data of the upcoming new scene. We build on Neural Radiance Fields (NeRF), which uses multi-layer perceptron to model the density and radiance field of a scene as the implicit function. While NeRF and its extensions have shown a powerful capability of rendering photo-realistic novel views in a single 3D scene, managing these growing 3D NeRF assets efficiently is a new scientific problem. Very few works focus on the efficient representation or continuous learning capability of multiple scenes, which is crucial for the practical applications of NeRF. To achieve these goals, our key idea is to represent multiple scenes as the linear combination of a cross-scene weight matrix and a set of scene-specific weight matrices generated from a global parameter generator. Furthermore, we propose an uncertain surface knowledge distillation strategy to transfer the radiance field knowledge of previous scenes to the new model. Representing multiple 3D scenes with such weight matrices significantly reduces memory requirements. At the same time, the uncertain surface distillation strategy greatly overcomes the catastrophic forgetting problem and maintains the photo-realistic rendering quality of previous scenes. Experiments show that the proposed approach achieves state-of-the-art rendering quality of continual learning NeRF on NeRF-Synthetic, LLFF, and TanksAndTemples datasets while preserving extra low storage cost.


\end{abstract}  
\section{Introduction}

Modeling and rendering photo-realistic novel views of real objects and scenes from images is a central task in computer vision and graphics, with various applications, such as virtual reality, autonomous driving, and robotics. Recently, Neural Radiance Fields (NeRF) \cite{n_vanilla_nerf} and its extensions, such as DVGO \cite{n_dvgo}, Mip-NeRF 360 \cite{mip-nerf360}, iNGP \cite{n_ingp}, and TensoRF \cite{n_tensorf}, have brought significant improvement by exploiting the representation power of neural networks. They represent a static scene with a multi-layer perceptron (MLP) by mapping the position and orientation information to the density and color domain. Volume rendering techniques are used along camera rays and integrate the output colors and densities from the MLP to achieve novel view synthesis. 

With the development of NeRF, how to efficiently managing these growing 3D NeRF assets is a new scientific problem. One of the naive approaches is to train a separate model for each scene. However, this approach is not scalable as the storage requirements and training time increase linearly with the addition of new scenes. In this paper, we aim to tackle this issue by designing a memory-efficient NeRF representation and developing a novel continual training framework for NeRF, so that 3D scenes can be kept learning, shared with a single network, and the network can be updated efficiently only with the training data of the upcoming new scene. As shown in Fig.\ref{teaser}, 3D NeRF assets can be efficiently stored and composited into arbitrary new scenes. The new coming 3D asset can be continually trained with the learned compact representation without the previous training data.

Like other modern deep neural networks \cite{resnet,transformer,gan}, NeRF is also susceptible to catastrophic forgetting: when adapted to perform new 3D scenes, they fail to generalize and cannot maintain their capability to accomplish previously learned 3D scenes. Due to their intrinsic differences, continual learning methods for previous discriminative models cannot be directly applied to reconstruction multiple NeRF scenes. First, there are severe conflicts between tasks under the NeRF setting. For discriminative models, it rarely happens that one image has different labels (appears in different tasks). However, for NeRF models, the same position and view direction have an extremely high probability of different density and radiance in various scenes. Second, it is well known that the intermediate convolutional layers in deep neural networks can provide generic features in classification tasks. Different from NeRF, other network architecture with different classification goals can easily reuse these features. Some methods have taken this idea and borrowed the convolutional module, the attention module, or more advanced neural network module to enhance the generalizability of NeRF for faster optimization, such as MVS-NeRF \cite{n_mvsnerf}, ContraNeRF \cite{n_contranerf}, CP-NeRF \cite{cp-nerf}, InsertNeRF \cite{insertnerf}, and IBRNet \cite{n_ibrnet}. However, the number of parameters for these methods is fixed, and the rendering quality decreases significantly as the number of scenes increases. Moreover, these methods are not designed for incremental 3D scene reconstruction scenarios. When new 3D scenes are added to the model for continual learning, the catastrophic forgetting problem will still occur for the previously trained scenes.

From another perspective, there has been some initial works \cite{n_meil_nerf,clnerf,il-nerf,icl-nerf} exploring NeRF with continual learning to reconstruction \textit{a single scene }with multiple sequences or appearance and geometry changes over an extended period of time. Different from them, we aim to mining the potential relationships in different scenes, design a memory-efficient representation for \textit{multiple scenes}, which can achieve photo-realistic rendering of both the upcoming new 3D scene and the previously learned scenes, given the constraint of only having access to the model trained on previous scenes without the previous data. One of the naive approaches to tackling the issue of catastrophic forgetting is to train a separate model for each task. However, this approach is not scalable as the storage requirements and training time increase linearly with the addition of new tasks.

In this paper, we introduce a generic continual learning framework, SCARF, that can memory-efficiently perform novel view synthesis across different scenes, be kept learning for new scenes, and update the network parameters only with the training data of the upcoming new scene. Our core idea is to factorize the parameters of multiple NeRF models into a set of scene-specific weight matrices for each scene and a cross-scene weight matrix that linearly combines with the scene-specific weight matrices. Instead of learning deterministic weight matrices for each scene, we learn to generate dynamic weight matrices from random noises using a global parameters generator. Multiple NeRF models share this parameter generator, and the parameter generator learns a generalizable prior for each hidden layer of NeRF, significantly reducing the number of parameters for reconstruction and rendering multiple scenes. Moreover, to introduce more flexibility for different scenes, the cross-scene weight matrix is multiplied by a small scene-specific coefficient matrix. For the continual learning process, the additional parameters required for the upcoming scene consist only of random noise and a small scene-specific coefficient matrix. Second, given a sequence of 3D scenes, knowledge is extracted from a previously trained model and distilled to the new model which was training for the new scenes, encouraging the new model to generate the same output as the previous model for previously learned scenes. We introduce an uncertain surface radiance field distillation strategy. We use this strategy to encourage the new model to generate the same output as the previous model of previous scenes efficiently and avoid distilling the low information entropy knowledge in previous trained scenes (blank parts inside and outside the surface of the 3D scene). SCARF is a plug-and-play approach. When the parameters of NeRF are generated by SCARF, the other pipeline (sampling, rendering, etc.) is same as NeRF, and can be combined with other NeRF extension work seamlessly. We evaluate our approach via continual learning on the NeRF-Synthetic \cite{n_vanilla_nerf}, TanksAndTemples \cite{exp_tt}, and LLFF \cite{exp_llff} datasets. Extensive ablation studies and comparisons with state-of-the-art models are conducted across diverse data domains. Qualitative and quantitative results are demonstrated to show the capability of our framework to learn new scenes with greatly overcoming the catastrophic forgetting of previously learned scenes.

To summarize, our contributions are as follows:
\begin{itemize}
    \item A novel memory-efficient and scalable representation of multiple NeRF, SCARF, compacts multiple scenes into a single small MLP whose parameters are generated by another tiny hypernetwork.
    \item A generic framework for continual learning NeRF based on SCARF that demonstrates how 3D scenes can continually learn and photo-realistic render for novel views. To the best of our knowledge, this is the first work trying to learn NeRF and render multiple 3D scenes continually.
    \item State-of-the-art performance in terms of rendering quality and storage for continual learning NeRF for multiple 3D scenes. In addition, SCARF can be easily plugged into other NeRF-based networks, and we show the further application of SCARF to efficiently organize, store, and incrementally learn the 3D implicit assets. 
\end{itemize}

\begin{figure*}[t]
\centering
\includegraphics[width=2.0\columnwidth]{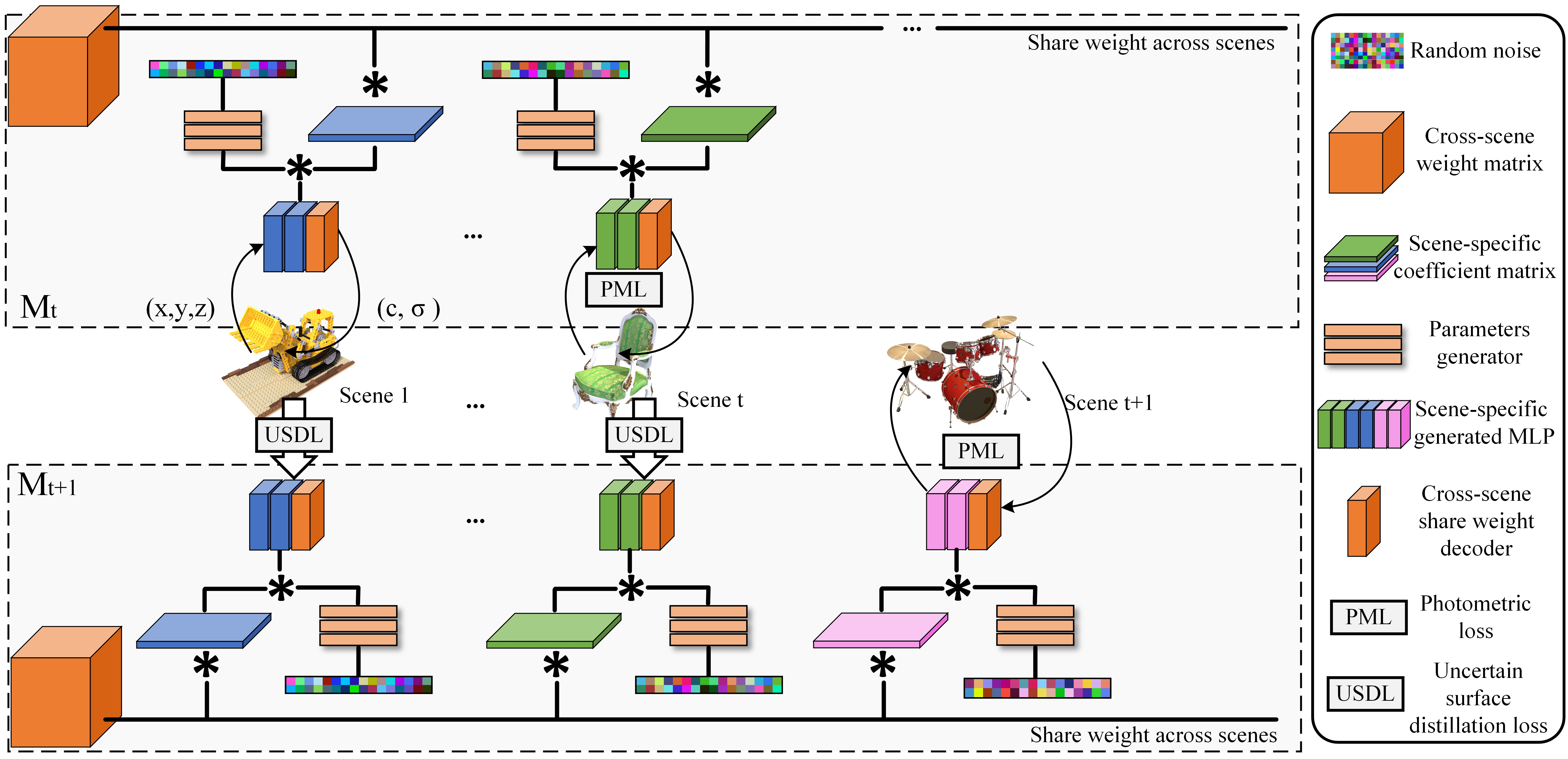}
\caption{With the proposed SCARF, given a sequence of 3D scenes, our method factorizes the MLP into a set of scene-specific weight matrices and a cross-scene weight matrix. A global parameter generator generates the scene-specific weight matrices, learning the generalizable features across scenes. Moreover, when a new 3D scene comes, additional parameters needed to introduce it into the network are only random noise and a coefficient matrix.}
\label{method}
\end{figure*}

\section{Related Work} 
\subsection{Neural Radiance Fields} Recently, NeRF \cite{n_vanilla_nerf} has gained significant attention and led to rapid breakthroughs in scene reconstruction and novel view synthesis. A growing number of subsequent NeRF extensions emerged, e.g., faster training \cite{n_ingp,n_autoint}, real-time inference \cite{n_mobilenerf,n_duplexnerf}, 3D scene editing \cite{n_sine,n_ripnerf}, etc. However, these works focus on applying it in a single scene and do not consider the efficient representation of multiple scenes. To this end, some works \cite{n_mvsnerf,n_contranerf,n_gm_nerf,n_ibrnet} have focused on generalizable NeRF, which aims to learn a single NeRF network on-the-fly by pre-training a set of scenes. However, they do not meet the needs of continuous learning task. First, finetune is still required on new scenes to guarantee high-quality rendering results, leading to catastrophic forgetting of previously learned scenes. In addition, the parameter number of the network in these methods is fixed, and the performance of the network degrades drastically with the increase of learned scenes. Another series of work related to us is the compression of NeRF. They use weight quantization \cite{n_quantizing_nerf}, low-rank approximation \cite{n_tensorf,n_ccnerf}, binarization \cite{n_binarynerf}, or knowledge distillation \cite{n_pvd} for additional optimization for compression NeRF models in a single scene. These compression-based works are orthogonal to our work; potentially, our multi-scene NeRF representation can be further compressed at the neural network parameter level by combining these methods, and we leave such combination as future work.

\subsection{Continual Learning} 
 Continual learning \cite{c_survey1,c_survey2} is a learning paradigm that aims to accomplish a sequence of new tasks while retaining the knowledge of previous tasks, given the constraint of only having access to a model trained on previous tasks without the previous data. To overcome the catastrophic forgetting problem, some works \cite{c_reg1,c_reg2} apply regularization loss during optimization to suppress network parameters that are important for past tasks. Other works apply replay-based approaches, such as experience replay \cite{c_replay1}, generative replay \cite{c_replay2}, and feature replay \cite{c_replay3}. Moreover, some works apply an architecture-based approach, such as parameter allocation \cite{c_arch1}, model decomposition \cite{c_arch2,c_arch3}, and modular networks \cite{c_arch4}. However, unlike these tasks, NeRF uses MLP to encode a scene's density and radiance field. There is a vast domain gap between different scenes, and the parameters cannot be reused efficiently with these methods.

 \subsection{Hyper- and meta- learning}
HyperNetworks \cite{hypernetworks} are neural networks designed to generate the weights of another network. Meta-learning, often described as "learning to learn," involves refining a learning algorithm through multiple episodes. MAML \cite{meta-maml} facilitates fast adaptation at test time by learning an initial model configuration using a gradient-based meta-learning approach. Building on MAML, CAVIA \cite{meta-cavia} adapts to new tasks by updating only a subset of input parameters rather than the entire network. Many studies integrate hyper-learning and meta-learning into computer graphics. MetaSDF \cite{meta-metasdf} optimizes weight initialization to quickly fit neural representations of signed distance fields. Matthew et al. \cite{meta-lll} apply meta-learning to initialize weights for fully-connected networks like NeRF. Metappearance \cite{meta-metaappearance} employs meta-learning for various appearance reproduction tasks. LoRA \cite{lora} enhances diffusion models by freezing pretrained weights and inserting trainable rank decomposition matrices, thus minimizing the number of trainable parameters for efficient fine-tuning. In contrast to these approaches, we investigate the potential relationships between NeRFs across different scenes and utilize hypernetworks to generate NeRF parameters for different scenes.

\subsection{Continual Learning for Neural Radiance Fields} 
There has been very few initial work exploring continual learning with NeRF. To achieve simultaneous localization and mapping (SLAM) with NeRF, some works \cite{imap,icl-nerf,slam-r-1,slam-r-2} propose to replay the keyframe to avoid network forgetting. Unlike our work, these works focus more on the real-time requirements and assume that the previous training data is always available (solving optimization for all keyframes in the bundle adjustment phase). MEIL-NeRF \cite{n_meil_nerf} propose to reconstruct a single scenes with multiple sequences. For each new sequence, it creates rays for a previously learned network using the ray generator network and jointly learns the generated and new rays. CLNeRF \cite{clnerf} proposes a replay-based strategy to reconstruct the scene with a sequence of multiple scans with appearance and geometry changes over an extended period of time. However, the above works all focused on a single scene, maintaining the partially learned regions unchanged by reducing the catastrophic forgetting problem through the continual learning strategy. Different from them, we focus on continual learning on different 3D scenes. Due to the vast domain gap between different scenes, these work are not applicable for efficient multiple NeRF reconstruction.

\section{Method}
The overall goal is to learn the model for a given sequence of scenes $ \{S_i\}_{i=1}^{N}$, assuming that the learning process has access to only one scene at a time and that training data for each scene is accessible only once. In this section, we start by briefly reviewing the vanilla NeRF pipeline. Then, we introduce SCARF. How to efficiently represent multiple NeRF models with a single small MLP whose parameters are generated by another tiny hypernetwork in SCARF is discussed. Finally, we introduce the framework for continual learning NeRF models based on SCARF.

\subsection{Preliminaries for vanilla Neural Radiance Fields}
In NeRF, a scene is represented by an implicit function $\mathbf{F}_\Theta$ that maps the spatial point $\mathbf{x}_i=(x, y, z)$ and view direction $\mathbf{d}_i=(\theta, \phi)$ into the density $\sigma_i$ and radiance $\mathbf{c}_i$, i.e.:
\begin{equation}
  [\sigma_i,\mathbf{c}_i] =\mathrm{F_{\Theta}}(\mathbf{x}_i, \mathbf{d}_i).
  \label{eq_vanilla_nerf_network}
\end{equation}

Given a ray $\mathbf{r}$ originating at $\mathbf{o}$ with direction $\mathbf{d}$, the spatial points $\mathbf{x}_i=\mathbf{o}+t_i\mathbf{d}$ are sequentially sampled along the ray. After querying $\mathbf{F}_\Theta$ and getting $\sigma_i$ and $\mathbf{c}_i$ for each point, the color of the pixel corresponding to the ray is then estimated by numerical quadrature:
\begin{equation}  \label{neural_rendering_equation}
    \hat{\mathbf{C}}(\mathbf{r}) = \sum_{i}^{N}T_i(1-\exp(-\sigma_i \delta_i))\mathbf{c}_i
\end{equation}
where $T_i = \exp(- \sum_{j=1}^{i-1} \sigma_j \delta_j)$, and $\delta_i = t_{i+1} - t_{i}$ is the distance between adjacent samples. However, aiming to model several scenes in sequence, vanilla NeRF suffers from catastrophic forgetting: when a new 3D scene is added, the vanilla NeRF model cannot perform photo-realistic rendering quality for previous scenes. Storing a separate model for each scene addresses catastrophic forgetting inefficiently, as each set of parameters is only helpful for one single scene.

\subsection{Representation of SCARF for multiple 3D scenes}
\label{method:3.2}
The radiance fields of different 3D scenes have significant domain gaps, so it is hard to adapt all parameters of a trained model to a coming up new scene, resulting in a degraded performance in either the previous scene or the new scene. Therefore, we propose to factorize the conventional MLP of the NeRF model into a set of scene-specific weight matrixes (SSWMs) and a cross-scene weight matrix (CSWM), which linearly combined with the SSWMs. Besides, to design a more memory-efficient network architecture for continual learning NeRF, we take inspiration from HyperNetworks \cite{hypernetworks} that use a single hypernetwork to generate the SSWMs for different scenes. Fig.\ref{method} illustrates the overall flow of SCARF. Now, we introduce the details of the factorization of NeRF.
\subsubsection{NeRF factorization}
Let the neural network be $\mathrm{F_{\Theta}}$ to represent $N$ 3D scenes, and $\mathrm{F_{\Theta}^i}$ to represent a specific scene $S_i$. Assuming the $\mathrm{F_{\Theta}^i}$ consists of the decoder $D$ and the $L$ layers' encoder $\{E^{i}_{l} \in \mathbb{R}^{c_{in} \times c_{out}}\}_{l=1}^{L}$ where $l$ denotes the index of layers, $c_{in}$ and $c_{out}$ are the input and output dim of the linear transformation for each layer, respectively. In our experiments, we define the first 9 layers of MLPs of vanilla NeRF as encoder  $\{E^{i}_{l}\}_{l=1}^{9}$, and the last 2 layers of MLPs as decoder $D$. Then $E^{i}_{l}$ is factorized into a SSWM $SM_{l}^{i} \in \mathbb{R}^{c_{in} \times K}$ and a CSWM $ CM_{l} \in \mathbb{R}^{K \times c_{out}}$, i.e.:
\begin{equation}  \label{nerf_factorization_equation}
    {E_{l}^i} =  {SM}_{l}^{i}* CM_{l}.
\end{equation}
To represent various 3D scenes, $\mathrm{F_{\Theta}}$ maintains different sets of SSWM $\{SM^{i}\}_{i=1}^{N}$, and for a specific scene $SM^{i}$ is composed of $L$ layers matrices $\{SM_l^{i}\}_{l=1}^{L}$ . To make the model parameter efficient, CSWM ${CM}=\{CM_l\}_{l=1}^{L}$ is shared across all scenes and learned the latent generalizable knowledge. By multiplying the CSWM ${CM}_{l}$ with the scene-specific SSWM ${SM}_{i}^{l}$ for scene $i$ layer $l$ respectively, the parameters of $L$ layers' encoder is generated. For the structural design of the decoder, we found that using one global cross-scene decoder $D$ is sufficient. Benefiting from CSWM, the encoder learns a generalizable high-dimensional unified scene representation, which allows us to transform the high-dimensional features into radiance and density fields with a single MLP decoder. 
\subsubsection{Parameters generator}
Although setting a smaller $K$ can largely reduce the number of parameters, we found that SSWMs are easily learned without constraints and overfitted, which leads to the degradation of rendering quality for multiple scenes. To address this problem, we take inspiration from HyperNetworks \cite{hypernetworks} that generates each layer $l$'s SSWM of each scene $i$ from random noise $z_l^i$ using a second meta neural network $G_l$ across scenes. Specifically,
\begin{equation}  \label{SSWM_paras_generator}
    SM_l^i =  {G}_{l}(z_l^i).
\end{equation}
Benefiting from the parameter generator, the learned SSWMs show better generalization capability when representing multiple scenes. Moreover, many more SSWMs of new scenes could be sampled from the vast parameter space by sampling different $z_l^i$ from some pre-defined distribution, which further reduced the number of parameters. In our work, we sampled the random noise from the normal distribution $N(0,1)$. While each SSWM is learned for a specific scene, the CSWM learns across scenes. $E_l^i$ can be seen as the linear combination of $CM_l$ with $SM_l^i$. There is a huge domain gap in different scenes of the NeRF model, and such a linear combination of CSWM and SSWM limits the capability to represent more high-frequency information. Therefore, we additionally introduce a scene-dependent linear transformation module. Maintaining a small learnable coefficient matrix $C_l^i \in \mathbb{R}^{K \times K}$ for each scene $i$ layer $l$ improves the flexibility of scene-dependent features. Specifically, 
\begin{equation}  \label{CSWM_paras_1}
    CM_{l}^{i} = C_{l}^{i} * CM_{l},
\end{equation}
and
\begin{equation}  \label{CSWM_paras_2}
    {E_{l}^i} =  {G}_{l}(z_l^i)* CM_{l}^{i} .
\end{equation}

\begin{figure*}[htbp]
\centering
\includegraphics[width=2.0\columnwidth]{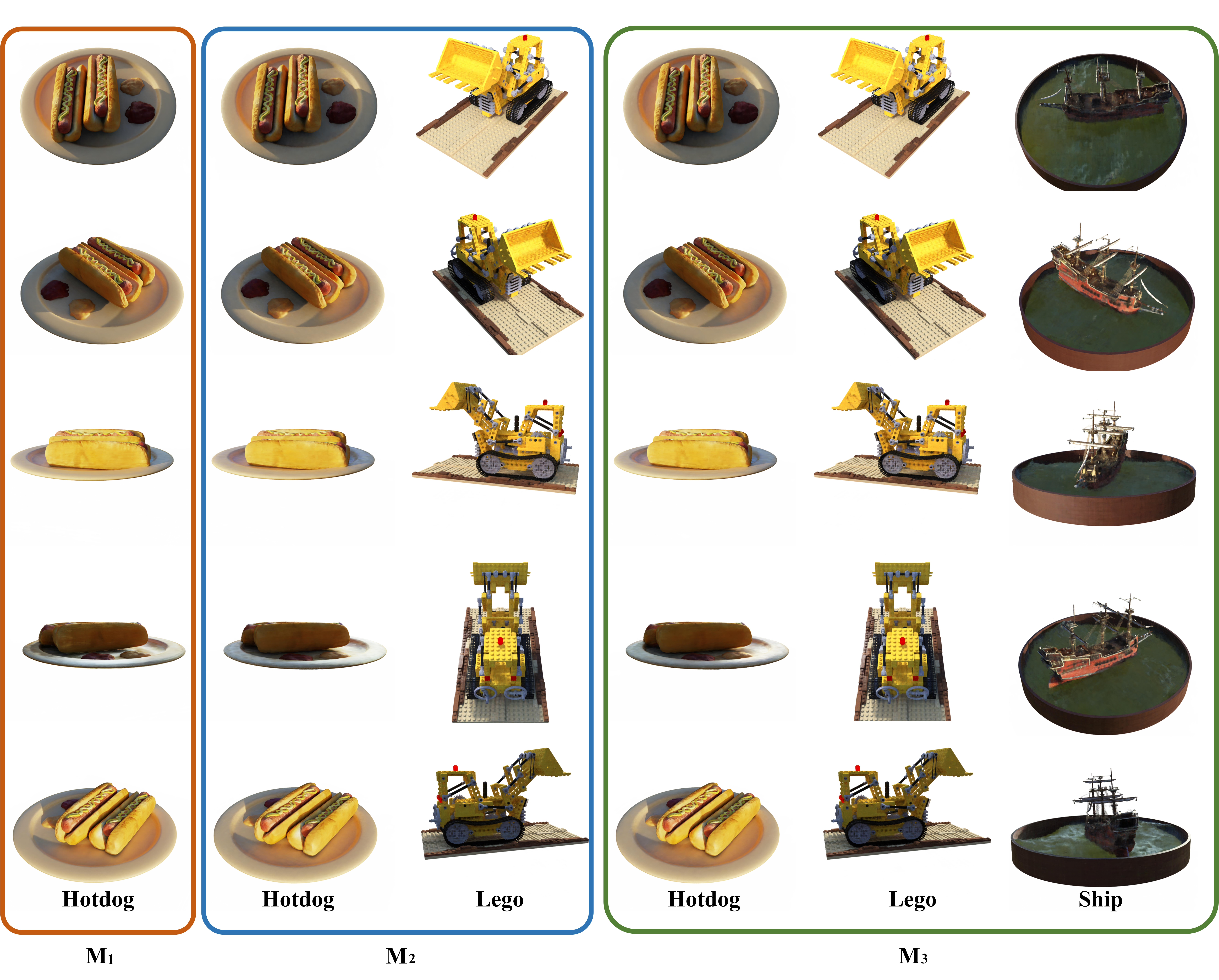}
\caption{Qualitative results of continual learning previous three scenes on the NeRF-Synthetic dataset.}
\label{exp_cl}
\end{figure*}

\begin{table}[htbp]
\resizebox{1.0\columnwidth}{!}{
\begin{tabular}{llllll}
\hline
Model    & Hotdog                       & Lego                       & Ship                          & Chair                        \\ \hline
$M_1$ (Hotdog)    & 36.61                    & --                         & --                            & --                             \\
$M_2 $ (Lego)    & ${36.45}$(  {$-$0.16})      & ${32.80}$                 & --                            & --                            \\
$M_3 $ (Ship)    & ${36.08}$(   {$-$0.37})     & ${32.49}$(   {$-$0.31})  & ${28.76}$                  & --                             \\ 
$M_4 $ (Chair)    & ${36.02}$(   {$-$0.06})    & ${32.41}$(   {$-$0.08})         & ${28.65}$(   {$-$0.11}) &${33.61}$                     \\ \hline
\end{tabular}}
\caption{Quantitative results of continual learning. We calculate the PSNR at different training stages for each scene. For the first column, the bracketed texts indicate the name of $n^{th}$ scene being trained for model $M_n$. For the other column, bracketed values indicate the difference compared to previous stage.}
\label{table-continual-learning-quantitative}
\end{table}

\subsection{Continual learning for SCARF}
\label{method:3.3}
With the help of the SCARF representation, multiple scenes can be compactly represented in a single neural network with scalability. Assuming that when the $t^{th}$ scene $S_{t}$ comes, the goal is to train a model $M_{t}$ based on $
M_{t-1}$ that could perform all scene $\{S_i\}_{i=1}^{t}$, while model $M_{t}$ is only restricted to the training data of the upcoming new scene $S_t$.
The model $M_{t-1}$ is composed of the following parameters:
a cross scene weight matrix $\{CM_l\}_{l=1}^{L}$ of $L$ layers,a cross-scene parameter generator $\{G_l\}_{l=1}^{L}$ of $L$ layers, $t-1$ scene specific coefficient matrices $\{C_l^i\}_{l=1,i=1}^{L,t-1}$, and $t-1$ scene specific random noises $\{z_l^i\}_{l=1,i=1}^{L,t-1}$. Note that $L$ indicates the layers of the neural network in the NeRF model, which is independent of the number of scenes. Given the new scene $S_t$, the additional parameters introduced only include the $L$ layers' random noise $\{z_l^{t}\}_{l=1}^{L}$ and the of $L$ layers' coefficient matrix $\{C_l^{t}\}_{l=1}^{L}$ for scene $t$. To overcome the catastrophic forgetting of the previous scenes $\{S_i\}_{i=1}^{t-1}$, it is necessary to learn the implicit representation of scene $S_t$ while distilling existing knowledge from $M_{t-1}$ to $M_{t}$.

\begin{figure*}[htbp]
\centering
\includegraphics[width=2.0\columnwidth]{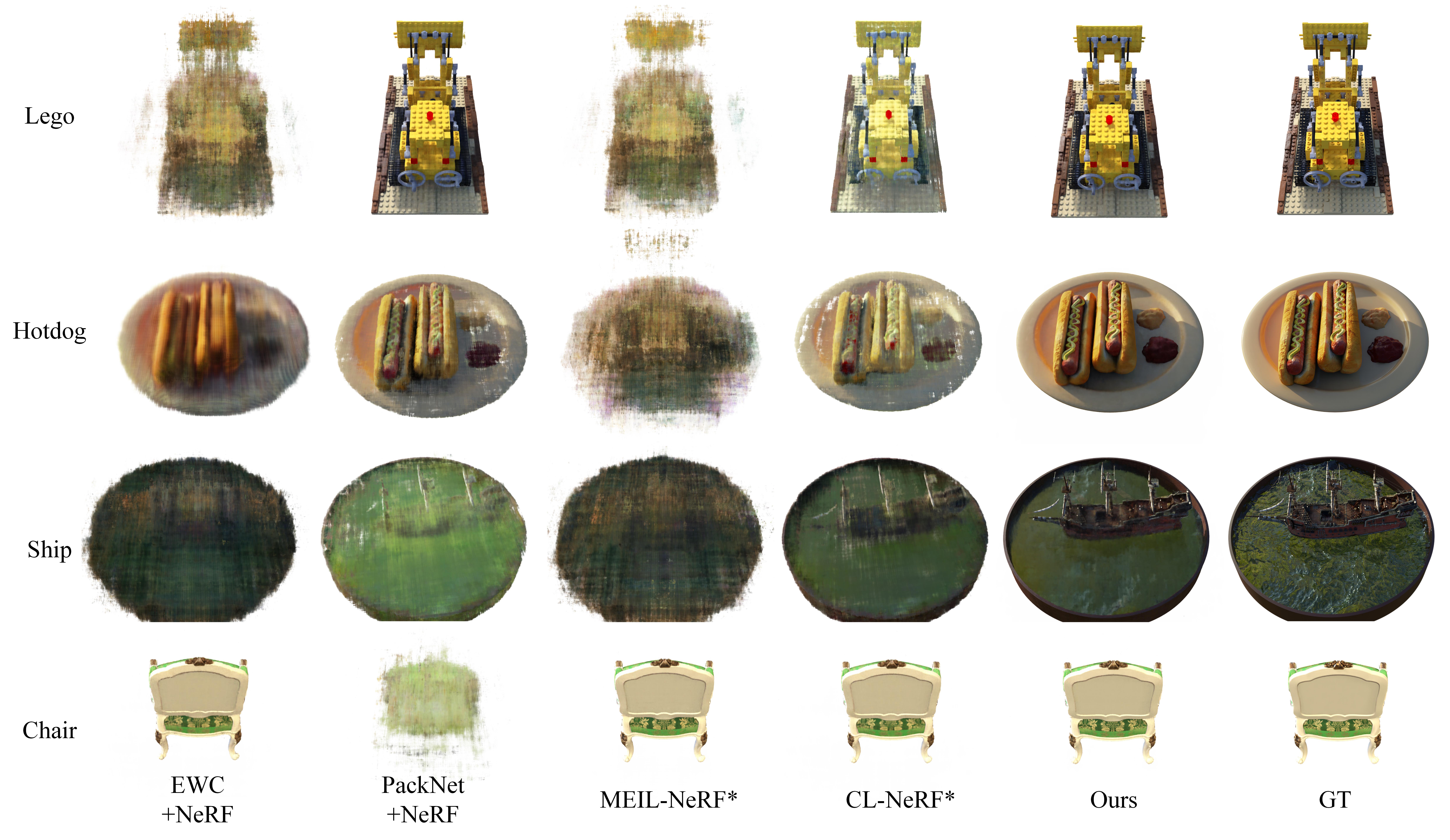}
\caption{Qualitative results of comparisons with some traditional continual learning methods combined with NeRF. "EWC+NeRF" is "continual learning multiple NeRF with Elastic Weight Consolidation \cite{c_reg1}", "PackNet+NeRF" is "continual learning multiple NeRF with PackNet \cite{packnet}", "MEIL-NeRF*" is "continual learning multiple NeRF with MEIL-NeRF \cite{n_meil_nerf}", and "CL-NeRF*" is "continual learning multiple NeRF with CL-NeRF".}
\label{exp_compare_nerf}
\end{figure*}

\subsubsection{Uncertain Surface Distillation}
We found that memory replay, a common solution for continual learning, on NeRF, i.e., direct distilling the knowledge from $M_{t-1}$ to $M_t$ for previous 3D scenes, as shown in Fig.\ref{exp_compare_nerf}, causes severe performance degradation.
A recent work \cite{n_pvd} introduces a Progressive Volume Distillation (PVD) strategy to achieve the conversions of a single scene between different NeRF architectures, e.g., from iNGP \cite{n_ingp} model to TensoRF \cite{n_tensorf} model. However, we found that since PVD is a multi-stage distillation, it cannot be applied to the continual learning of multiple 3D scenes. Furthermore, for memory-compact multi-scene representation, PVD is overfitted in a single scene and thus fails to maintain high-quality rendering results of multiple scenes.
So, we introduce a single-stage NeRF knowledge distillation method for continual learning NeRF to distill the knowledge of previous scenes from $M_{t-1}$ to $M_{t}$, called Uncertain Surface radiance field Distillation (USD). PVD has shown that distilling the radiance field (radiance and density) and the rendered RGB pixels preserves better rendering quality in NeRF than directly distilling the network parameters. We similarly distill the knowledge from the model $M_{t-1}$ in the radiance field level and RGB pixel level, i.e.:
\begin{equation}  \label{distillation_equ_cd}
    \mathcal{L}_{\mathbf{c\sigma}}^{i} = ||\hat{\mathbf{c}_{M_{t}}^i}-\mathbf{c}_{M_{t-1}}^i||_2 +\alpha||\hat{\mathbf{\sigma}_{M_{t}}^i}-\mathbf{\sigma}_{M_{t-1}}^i||_2,
\end{equation}

\begin{equation}  \label{distillation_equ_d}
    \mathcal{L}_{C(r)}^i = ||\hat{\mathbf{C(r)}_{M_{t}}^i}-\mathbf{C(r)}_{M_{t-1}}^i||_2 .
\end{equation} 
$\mathbf{c}_{M_{t-1}}^i$ and $\mathbf{\sigma}_{M_{t-1}}^i$ are the density and radiance random sampled from $M_{t-1}$ in scene $S_i$, while $\hat{\mathbf{c}_{M_{t}}^i}$ and $\hat{\mathbf{\sigma}_{M_{t}}^i}$ are corresponding the density and radiance sampled from $M_{t}$ in scene $S_i$. $\alpha$ is a hyperparameter balancing the weights of two regular terms. Moreover, we constrain the rendered pixel (using Eq.\ref{neural_rendering_equation}) $\hat{\mathbf{C(r)}_{M_{t}}^i}$ from model $M_{t}$ close to $\mathbf{C(r)}_{M_{t-1}}^i$ from model $M_{t-1}$. Unlike PVD, USD achieves the knowledge distillation of the NeRF model in a single stage. To this end, taking inspiration from multi-task learning \cite{mtl_loss}, we introduce the uncertain loss and use the learnable self-balancing parameters to distill both the density-radiance field and the RGB pixel simultaneously. Two learnable parameters $\beta_1$ and $\beta_2$ are introduced to weigh the uncertainty and balance the weight between these two loss functions. Due to the large domain gap between $\mathcal{L}_{\mathbf{c\sigma}}^{i}$ and $L_{C(r)}^i$, we found that this is very necessary. Moreover, we observe that since the density field is sparse, the random sampling distillation strategy leads to sampling a large number of blank regions (positions where the corresponding density tends to zero), making it unable to distill previous knowledge into new models efficiently. Therefore, for the loss $\mathcal{L}_{\mathbf{c\sigma}}^{i}$, we pre-extract an explicit density occupancy grid for each learned scene and only sample the radiance field and on the surface, which the density $\sigma_{i}$ greater than a threshold $\tau$. Specifically,

\begin{equation}  \label{auto_balance}
    \mathcal{L}_{dis}^i = \beta_1\mathbb{I}(\sigma_i > \tau) L_{\mathbf{c\sigma}}^{i}  + \beta_2  \mathcal{L}_{C(r)}^i + \log{\beta_1 \beta_2},
\end{equation} 
in which $\mathbb{I}(.)$ is the indicator function.
\subsubsection{Joint Learning}
While distilling knowledge from $M_{t-1}$, the network simultaneously learns new scene $S_t$. We follow vanilla NeRF, predicting the radiance field for new scenes and utilizing the volume rendering techniques supervised on RGB pixel space, i.e.:
\begin{equation}  \label{vanilla_nerf_eq}
    \mathcal{L}_{C(r)}^t = ||\hat{\mathbf{C(r)}_{M_{t}}^t}-\mathbf{C(r)}^t||_2 ,
\end{equation} 
while $\hat{\mathbf{C(r)}_{M_{t}}^t}$ is the predicted pixel RGB color from $M_t$ of the ray $r$ in scene $t$ and $\mathbf{C(r)}^t$ is the counterpart ground truth pixel RGB color. Our complete loss function during continual learning is as follows:
\begin{equation}
\mathcal{L} = \sum_{i=1}^{t-1}{\mathcal{L}_{dis}^i} + \gamma \mathcal{L}_{C(r)}^t,
\label{total_loss}
\end{equation}
and $\gamma$ is a hyperparameter balancing the weights of two regular terms. The former item distills knowledge for the previous $t-1$ scenes, and the latter item learns new coming-up scene $t$, achieving continual learning NeRF.

 \begin{table}[htbp]
\centering
\begin{tabular}{lcccc}
\hline
    \multirow{2}{*}{Method} & \multicolumn{2}{c|}{LLFF}                        & \multicolumn{2}{c}{TanksAndTemple}             \\ \cline{2-5} 
                        & PSNR           & SSIM           & PSNR           & SSIM           \\ \hline
EWC + NeRF                 & 14.90          & 0.412   & 15.64          & 0.420                 \\ 
PackNet + NeRF              & 16.67          & 0.551                         & 16.71          & 0.547                       \\ 
SLE + NeRF                    & 16.87           &0.544                      & 19.31             &0.608                  \\
MEIL-NeRF$^{*}$              & 17.27          & 0.571                         & 17.98          & 0.580                       \\ 
CLNeRF$^{*}$              & 21.67          & 0.659                         & 21.30          & 0.640                       \\ 

Ours              & \textbf{26.44} & \textbf{0.808}               & \textbf{26.78} & \textbf{0.892}             \\ \hline

\end{tabular}
\caption{Comparison of the quantitative results of common continual learning methods combined with NeRF.}
\label{compare-cl}
\end{table}

\begin{figure}[htbp]
\centering
\includegraphics[width=1.0\columnwidth]{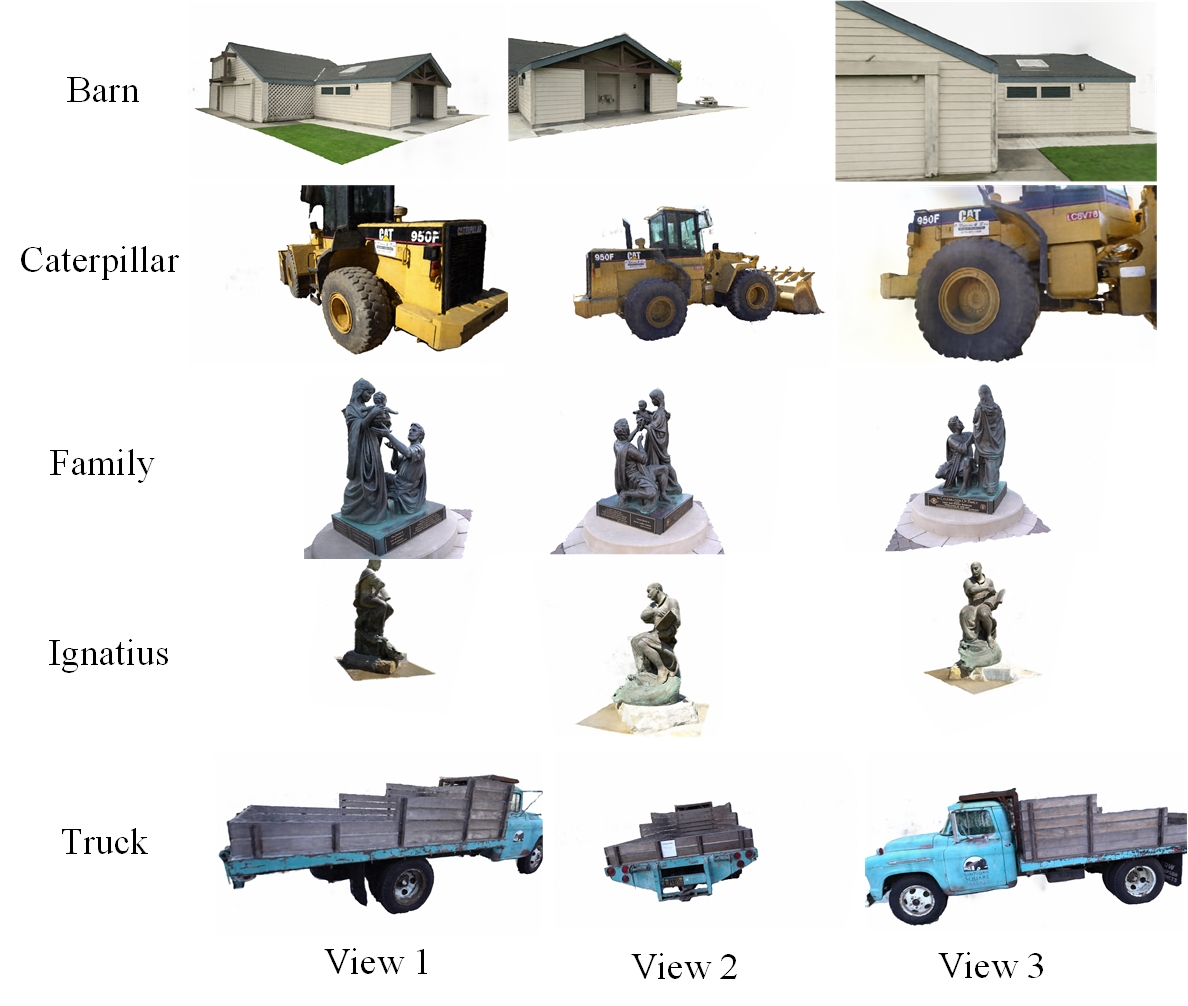}
\caption{Qualitative results of continual learning of five scenes on TanksAndTemples dataset.}
\label{sup_vis_tt}
\end{figure}
\begin{figure}[htbp]
\centering
\includegraphics[width=1.0\columnwidth]{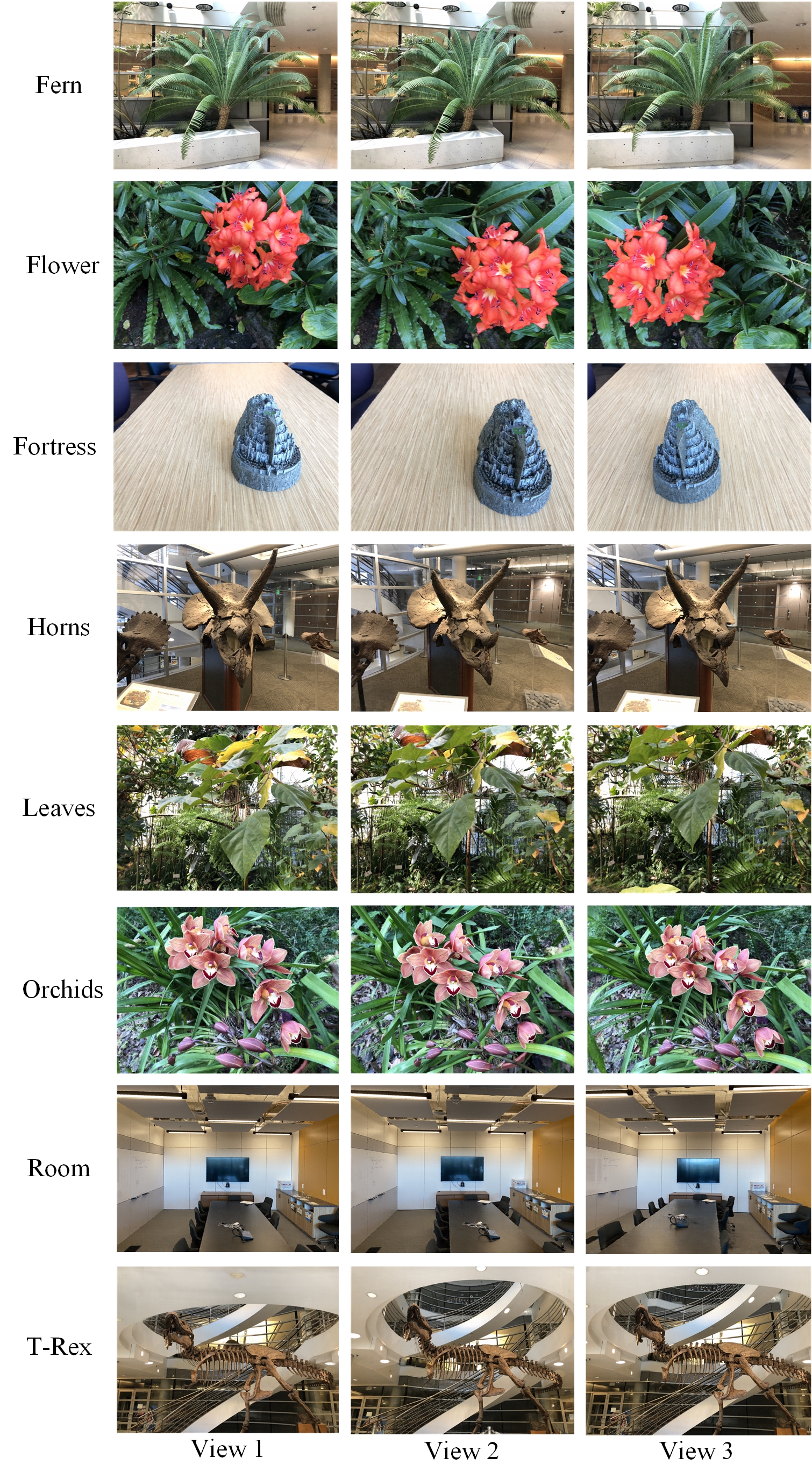}
\caption{Qualitative results of continual learning of eight scenes on LLFF dataset.}
\label{sup_vis_llff}
\end{figure}
\section{Experiments}
\subsection{Implementation Details}
\subsubsection{Datasets} The datasets utilized in this paper involve NeRF-Synthetic dataset, \cite{n_vanilla_nerf} forward-facing dataset (LLFF) \cite{exp_llff}, and TanksAndTemple dataset \cite{exp_tt}. The scenes within each dataset are input into the model in sequence for continual learning. The ground truth label of previous scenes is unavailable when training the upcoming new scene.
\subsubsection{Network Architecture}
We briefly discussed our implementation; please refer to the supplementary materials for more details. For the architecture of the NeRF model, we keep consistent with the original settings \cite{n_vanilla_nerf} as much as possible. We use a fully-connected network with an eight-layer encoder and a two-layer decoder with a ReLU \cite{exp_relu} layer, and the encoders are decomposed into SSWMs and CSWM.

\begin{table*}[htbp]
\centering
\begin{tabular}{crrrrcccc}
\hline
\multirow{2}{*}{Method} & \multicolumn{4}{c}{Size (MB) $\downarrow$ } & \multicolumn{2}{c}{Capability}   & \multicolumn{2}{c}{NeRF-Synthetic}                   \\ \cline{2-9} 
                               &1 scenes &3 scenes &5 scenes &8 scenes         &CP             & CL           & PSNR$\uparrow$   & SSIM$\uparrow$                      \\ \hline
NeRF \cite{n_vanilla_nerf} &\textbf{2.50} &7.50 &12.50 &20.00 &\XSolidBrush &\XSolidBrush &31.01 &0.947 \\         
CC-NeRF-HY \cite{n_ccnerf}      &88.00 &264.00&440.00 &704.00       &\Checkmark     &\XSolidBrush  & 32.37          & 0.955                                                    \\ 
CC-NeRF-CP \cite{n_ccnerf}      &4.42   &13.26   &22.1   &35.36    &\Checkmark     &\XSolidBrush  & 30.55          & 0.935              \\
DVGO \cite{n_dvgo}              &105.90&317.70&529.50  &847.20     &\XSolidBrush &\XSolidBrush  & \textbf{34.90}          & 0.899                     \\ 
VQ-Plenoxels \cite{n_vqnerf}     &13.70&41.10&68.50    &109.60   &\Checkmark&\XSolidBrush  & 31.53          & 0.956 \\                       
VQ-TensoRF \cite{n_vqnerf}       &3.60 &10.80 &18.00  &28.80      &\Checkmark&\XSolidBrush &32.86   &\textbf{0.960} \\  \hline
SCARF with HI (Ours)      &3.21&\textbf{3.23}&\textbf{3.25} &\textbf{3.28}    &\Checkmark&\XSolidBrush  & 31.57 & 0.957  \\
SCARF (Ours)      &3.21&\textbf{3.23}&\textbf{3.25} &\textbf{3.28}     &\Checkmark&\Checkmark  & 30.94 & 0.945          \\\hline

\end{tabular}
\caption{Comparison with recent methods related to composition and compression NeRF. Our method achieves comparable rendering results with the most minimal model size for multiple scenes while enabling compression (CP) and continual learning (CL). Metrics (PSNR and SSIM) are averaged over the eight scenes. "SCARF with HI" indicates that the historical images of learned scenes is always accessible, which demonstrates the compression capability of our model.}
\label{compare-cc}
\end{table*}

\subsubsection{Training and Distilling Details}
 We use the Adam Optimizer \cite{exp_adam} with initial learning rates of $5e-4$ for the learnable matrices (contains CSWM, coefficient matrices) and $1e-4$ for the parameters generator. We use the hyperparameters $\alpha = 3$ and $\gamma = 0.2$ at the training stage to balance the different losses. The learnable parameters $\beta_1$ and $\beta_2$ are initialized to $0.045$ and $0.06$ with the learning rate of $8e-5$ to weigh the uncertainty, respectively. Please check the supplementary materials for more details.
\subsection{Performance and Efficiency}

\subsubsection{Continual learning Results}
We are the first work to continually learn multiple scenes of NeRF, to the best of our knowledge. So, our work focuses on whether continual learning of NeRF can maintain performance and whether the model parameters keep memory efficiency with the increasing number of coming scenes. We first evaluate our model on the NeRF-Synthetic dataset. The scenes (Hotdog, Lego, Ship, and Chair) are input into the model in sequence, and the ground truth label of previous scenes is unavailable when training the new scene. Tab.\ref{table-continual-learning-quantitative} shows the quantitative results of continual learning NeRF for 4 scenes in the NeRF-Synthetic dataset with PSNR and SSIM. Fig.\ref{exp_cl} shows the corresponding qualitative results of the previous 3 scenes. We can see that high quality results can be achieved for previously learned scenes.
\subsubsection{Upper limit of the learned scenes} 

 \begin{table}[htbp]
\centering
\resizebox{1.0\columnwidth}{!}{
\begin{tabular}{lcccccc}
\hline
    \multirow{2}{*}{Method} & \multicolumn{2}{c}{LLFF}                        & \multicolumn{2}{c}{T\& T}   & \multicolumn{2}{c}{NeRF-Synthetic}            \\ \cline{2-7} 
                        & PSNR           & SSIM           & PSNR           & SSIM   & PSNR           & SSIM          \\ \hline
CLSD                 & 26.44          & 0.808   & 26.78          & 0.892   &30.94 &0.945                 \\ \hline
CLCD              & 26.21          & 0.801   & 25.97          & 0.889   &30.92 &0.944                      \\ \hline

\end{tabular}}

\caption{Experiments on the impact of the number of learned scenes on rendering quality. "CLSD" refers to "Continuous Learning in Single Dataset ," which involves continual learning within a single dataset—specifically, 5 scenes for the TanksAndTemples (T\&T) dataset, and 8 scenes each for the LLFF and NeRF-Synthetic datasets. Conversely, "CLCD" stands for " Continuous Learning Cross Dataset", where learning occurs across the aforementioned datasets, encompassing a total of 21 scenes.}

\label{up-1}

\end{table}

\begin{table}[htbp]
\centering
\begin{tabular}{lccc}
\hline
                  & PSNR$\uparrow$           & SSIM$\uparrow$     & Size (MB)$\downarrow$        \\ \hline
Group 1 (8 scenes)       & 30.94          & 0.945        & 3.28                \\
Group 2 (16 scenes)       & 30.89          & 0.941       & 3.36                     \\
Group 3 (24 scenes)       & 30.87          & 0.939       & 3.44                    \\
Group 4 (30 scenes)       & 30.87          & 0.938       & 3.52                  \\
Group 5 (36 scenes)     & 30.64          & 0.931         & 3.60                  \\
Group 6 (42 scenes)       & 30.67          & 0.931       & 3.68                  \\
Group 7 (48 scenes) & 30.65          & 0.930             & 3.76            \\\hline

\end{tabular}

\caption{Experiments on the impact of the number of learned scenes on rendering quality with data augmentation for NeRF-Synthetic dataset. For each experimental group, we generated 8 novel scenes from the original dataset and performed continual learning on these augmented scenes.}
\label{up-2}
\end{table}

As our work is the first to focus on continuous NeRF learning across multiple scenes, there is currently no established baseline for quantitatively analyzing how the number of learned scenes affects the rendering quality of previously learned scenes. We conduct two experiments to quantitatively explore the upper limit of learned scenes: continuous learning across datasets and repeated continuous learning within a single dataset using data augmentation. For continuous learning across datasets, we sequentially trained on the LLFF dataset (containing 8 scenes), the TanksAndTemples dataset (containing 5 scenes), and the NeRF-Synthetic dataset (containing 8 scenes), resulting in a total of 21 scenes. As a baseline for comparison, we also provide results of continuous learning within a single dataset. Tab.\ref{up-1} demonstrates that as the number of scenes increases from 5 or 8 to 21, the rendering quality remains comparable. Additionally, we perform data augmentation on the NeRF-Synthetic dataset to generate multiple group of datasets with domain gaps but that should achieve similar rendering quality (PSNR) after training. This strategy allows us to continuous learning these augmented datasets group by group and observe changes in rendering quality, thus quantifying the robustness of SCARF to the increasing number of learning scenes. For each scene in the NeRF-Synthetic dataset, we apply the following transformations sequentially: flip the X-axis, flip the Y-axis, flip the Z-axis, swap the X-axis and Y-axis, swap the Y-axis and Z-axis, and swap the X-axis and Z-axis. Consequently, 6 new scenes with domain gaps are generated for each scene, producing 6 groups totaling 48 scenes. We learn each group of scenes sequentially and calculate the average metrics for all learned scenes. As shown in Tab.\ref{up-2}, even with 48 learned scenes, each NeRF still demonstrates well rendering performance. This success is attributed to the disentanglement of scene-specific matrices and the cross-scene matrix, where the cross-scene matrix has learned generalizable features capable of handling various domains.
 \subsubsection{Comparisons with baseline}

We compare SCARF to the following baseline models, which combine NeRF with the common continual learning methods:
(a) Elastic Weight Consolidation (EWC) \cite{c_reg1} + NeRF: EWC is a widely-used regularization-based continual learning method.
(b) PackNet \cite{packnet} + NeRF: PackNet is a parameter isolation continual learning method that pruning less important parameters for past tasks.
(c) SLE + NeRF: Since vanilla NeRF cannot represent multiple scenes, we integrate Scene-Level Learnable Embeddings (SLE) into each NeRF model. The network architecture is designed similarly to NeRF-W \cite{nerf_w}, and we optimize the SLE using Generative Latent Optimization (GLO) techniques akin to those used in NeRF-W.
(d) MEIL-NeRF$^{*}$ : MEIL-NeRF \cite{n_meil_nerf} is designed for continual learning multiple sequences in a single scene. And it cannot apply for multiple scenes reconstruciton scenario. For fair comparsion, we improve the MEIL-NeRF with a scene-specific learnable latent code as an additional input, which is called MEIL-NeRF$^{*}$.
(e) CLNeRF$^{*}$ : CLNeRF \cite{clnerf} is designed for continually learning multiple sequences with appearance and geometry changes over an extended perios in a single scene. And it also cannot apply to multiple scenes reconstruction scenarios. For fair comparsion, we also improve the CLNeRF with a scene-specific learnable latent code as an additional input, which is called CLNeRF$^{*}$.

Tab.\ref{compare-cl} shows the results of the quantitative analyses on LLFF and TanksAndTemple datasets, while Fig.\ref{exp_compare_nerf} shows qualitative results of continual learning in four scenes of NeRF-Synthetic dataset. In contrast to traditional methods of knowledge distillation, we can learn new scenes while preserving the rendering quality of previous learned scenes. Besides, Fig. \ref{sup_vis_tt} and Fig. \ref{sup_vis_llff} show qualitative results of continual learning on TanksAndTemple (5 scenes) and LLFF (8 scenes) datasets, respectively.

\subsubsection{Comparisons with composable and compression NeRF} 

We also compare our method with some recent works on composable and compression NeRF in Tab.\ref{compare-cc}. We focus on the continual learning capability to facilitate practical applications, but not boosting the rendering quality over the previous state-of-the-art. Although the rendering performance of the proposed method is not the best, the memory-efficient model design with the extra continual learning capability is unique and enables various applications.
\begin{table}[htb]
\resizebox{0.95\columnwidth}{!}{
\begin{tabular}{cccc}
\hline
                  & PSNR$\uparrow$           & SSIM$\uparrow$           & Size (MB)$\downarrow$      \\ \hline
Dim of $z$ = 8       & 29.63          & 0.940          & 2.12                    \\
Dim of $z$ = 32       & 30.91          & 0.942          & 5.82                    \\\hline
$K$ = 15       & 29.23          & 0.939          & 2.51                   \\
$K$ = 27       & 30.98          & 0.943          & 4.42                    \\\hline
w/o coefficient matrix     & 28.75          & 0.938          & 3.26                    \\
w/o parameter generator       & 29.62          & 0.941          & 8.35                    \\
w/o $\mathcal{L}_{\mathbf{c\sigma}}$ & 28.45          & 0.937          & 3.28                 \\
w/o $\mathcal{L}_{C(r)}$    & 24.61          & 0.910          & 3.28                    \\
w/o uncertain weight   & 26.15          & 0.916          & 3.28 \\
w/o surface distillation    & 27.29          & 0.929          & 3.28                    \\\hline
w/ all            & {30.94} & {0.945} & 3.28  \\ \hline
\end{tabular}
}
\caption{Ablation studies of our method. PSNR and SSIM metrics are averaged over the eight scenes from the NeRF-Synthetic dataset, which are continually learned.}
\label{ablation_study}
\end{table}
\subsection{Ablation studies}
\subsubsection{What contributes to our rendering quality}
Our ablation studies validate the algorithm's design choice on the NeRF-Synthetic dataset in Tab.\ref{ablation_study}. We implement continual learning of 8 scenes on the NeRF-Synthetic dataset. Rows 1-4 show our choice of dim of random noise $z$ and hyperparameter $K$. Only using 8 dims of the random noise $z$ reduces performance, but increasing the number of the dims of random noise $z$ to 32 does not improve performance. A larger dim coefficient matrix $C$ can provide greater flexibility for modeling. However, larger $K$ increases network parameters. So, we choose $K=21$ as a trade-off. Row 5 demonstrates the performance will drop sharply without the coefficient matrix $C$. What's more, we find that parameter generator $G(.)$ not only reduces the number of parameters in the model but also improves the quality of the rendering in row 6. In rows 7-10, we also take continual learning without using the loss of $\mathcal{L}_{\mathbf{c\sigma}}$, $\mathcal{L}_{C(r)}$, uncertain learnable weight, or surface occupy grid distillation. The experimental results show that pixel-level distillation (with $\mathcal{L}_{C(r)}$) is the most crucial factor, as it efficiently optimizes multiple 3D sampling points along a ray like NeRF. Additionally, surface distillation of radiance and density at the 3D level is essential for further improving rendering quality.
\subsubsection{Ablations on the order of the scenes of continual learning}
\begin{figure}[htbp]
\centering
\includegraphics[width=0.90\columnwidth]{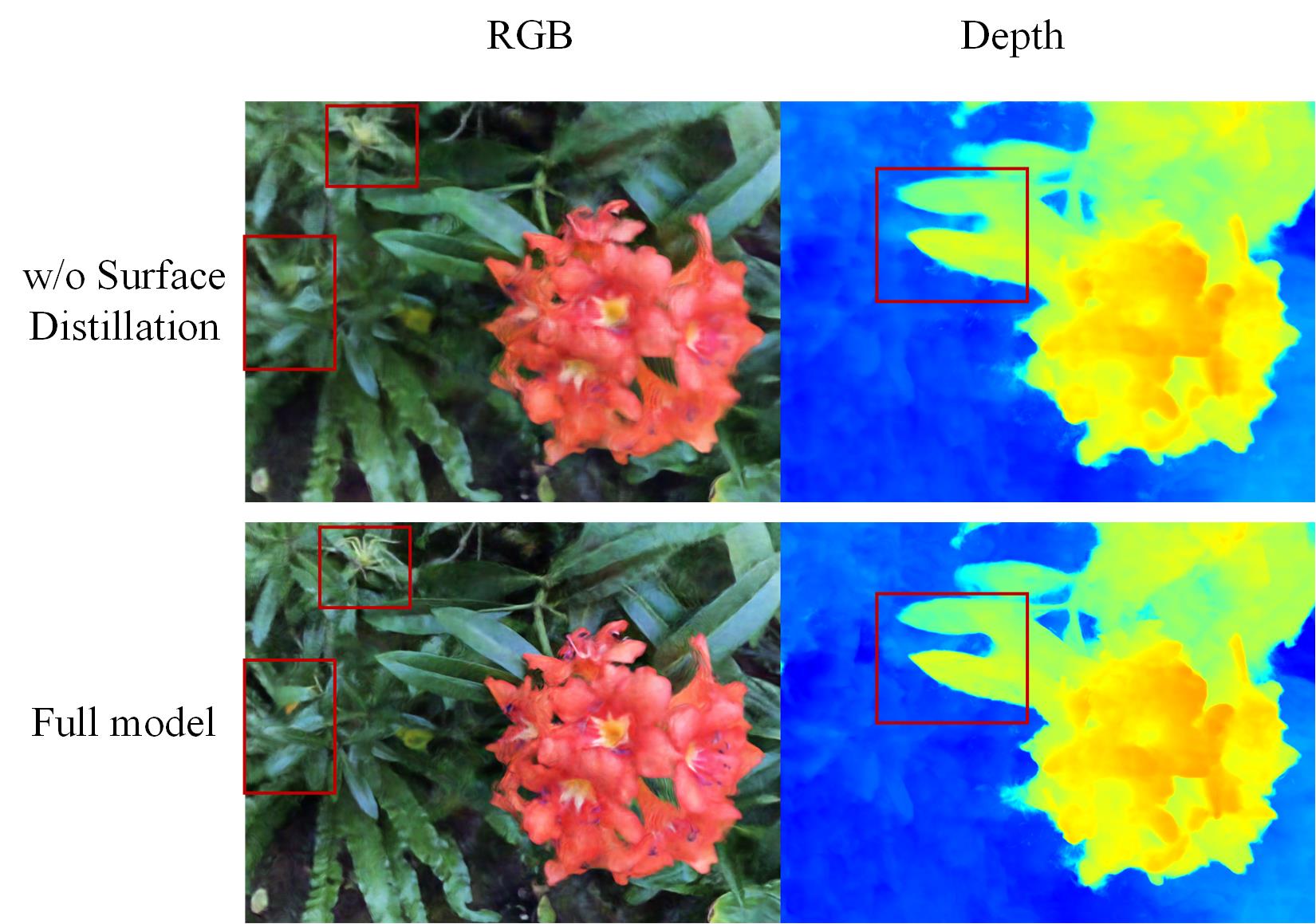}
\caption{Ablation studies on the USD strategy. The results indicate that the USD improves the RGB and depth quality.}
\label{exp_ablation_skd}
\end{figure}

Furthermore, we demonstrate the ablation studies on the order of continual learning. As shown in Tab.\ref{ablation_study_switch}, conducting multiple sets of experiments that swapped the order in which scenes were learned, we found that the rendering quality (PSNR) of the latest scene would be better. However, after training about two scenes, the rendering quality of the previous scenes stabilizes. Moreover, as shown in Fig.\ref{exp_ablation_skd}, the USD strategy made knowledge distillation more efficient and improved the RGB and depth accuracy of previous scenes. 

\begin{table}[htb]
\resizebox{1.0\columnwidth}{!}{
\begin{tabular}{llllll}
\hline
Stage       & Hotdog                       & Lego                       & Ship                          & Chair                        \\ \hline
H$\rightarrow$L$\rightarrow$S$\rightarrow$C    & $36.02$                     & $32.41$                          & $28.65$                            &$33.61$                  \\
L$\rightarrow$S$\rightarrow$C$\rightarrow$H     & $36.59^{(+0.57)}$                    & $32.40^{(-0.01)}$                          & $28.57^{(-0.08)}$                           &$33.39^{(-0.22)}$                  \\
S$\rightarrow$C$\rightarrow$H$\rightarrow$L     & $36.46^{(+0.44)}$                    &$32.81^{(+0.40)}$                        & $28.56^{(-0.09)}$                            &$33.22^{(-0.39)}$                  \\
C$\rightarrow$H$\rightarrow$L$\rightarrow$S     & $36.47^{(+0.45)}$                 & $32.67^{(+0.26)}$                          & $28.79^{(+0.12)}$                            &$32.18^{(-0.43)}$                 \\\hline
\end{tabular}}
\caption{Ablation studies on the order in which each scene is learned earlier. "H", "L", "S", and "C" are scenes of "Hotdog", "Lego", "Ship", and "Chair", respectively. The first column of the table shows the continual learning order.}
\label{ablation_study_switch}
\end{table}

\subsubsection{Ablations on using vanilla fully-connected decoder}
\begin{table}[htb]
\resizebox{1.0\columnwidth}{!}{
\begin{tabular}{cccc}
\hline
                  & PSNR$\uparrow$           & SSIM$\uparrow$ &Size (MB)               \\ \hline
Generate decoder for every scene  &30.92 & 0.943   & 2.81 \\
With  per-scene decoders  &30.97 & 0.946   & 7.17 \\
With a global decoder             & 30.95 & 0.944 & 3.28  \\ \hline

\end{tabular}
}
\caption{Ablation studies on the design choices on the decoder. Metrics (PSNR and SSIM) are averaged over the eight scenes on the NeRF-Synthetic dataset, which are continually learned.}
\label{sup_decoder}
\end{table}
At first, we tried to generate the decoder's parameters from another hypernetwork, just as we did with the encoder. As shown in Tab.\ref{sup_decoder}, we find that such a change does not improve the quality of the rendering. We considered that since the parameters of the encoder were generated through a global parameter generation network, in which the encoded high-dimensional features are generalizable, the use of a global decoder across scenes is sufficient. For simplicity yet efficient purposes, we use a global decoder.

\section{Conclusions}
In this work, we present SCARF, a generic continual learning framework that can perform novel view synthesis across multiple 3D scenes memory-efficiently. With SCARF, 3D NeRF assets can be efficiently stored and composited into arbitrary new scenes, while the new coming 3D asset can be continually trained with the learned compact representation without the previous training data. The key idea is to compact multiple scenes into a single small MLP whose parameters are generated by another tiny hypernetwork. Furthermore, we introduce the USD to distill the knowledge from previously learned NeRF. Experiments demonstrated that our approach achieves photo-realistic rendering results with the extra continual learning capability and extremely low memory cost.

Our method also has some limitations inherited from continual learning. For example, the catastrophic forgetting problem can only be greatly overcome and not directly addressed, which will further affect the rendering quality of NeRF for previously learned scenes. How to continually learn a large number of tasks remains an open problem.

\section{Acknowledgment}
This paper is supported by National Natural Science Foundation of China (No. 62072020); the Leading Talents in Innovation and Entrepreneurship of Qingdao, China (19-3-2-21-zhc); Open Project Program of State Key Laboratory of Virtual Reality Technology and Systems, Beihang University (No.VRLAB2024A**); and National Natural Science Foundation of China (No. 62201018).

\bibliographystyle{eg-alpha-doi} 
\bibliography{egbibsample}       

\newcommand{\etalchar}[1]{$^{#1}$}
\begin{thebibliography}{\uppercase{GPAM{\etalchar{*}}14}}

\bibitem[BDH{\etalchar{*}}23]{insertnerf}
\textsc{Bao Y., Ding T., Huo J., Li W., Li Y., Gao Y.}:
\newblock Insertnerf: Instilling generalizability into nerf with hypernet modules.
\newblock \emph{arXiv preprint arXiv:2308.13897} (2023).

\bibitem[Ben22]{c_reg2}
\textsc{Benzing F.}:
\newblock Unifying importance based regularisation methods for continual learning.
\newblock In \emph{International Conference on Artificial Intelligence and Statistics} (2022), PMLR, pp.~2372--2396.

\bibitem[BMV{\etalchar{*}}22]{mip-nerf360}
\textsc{Barron J.~T., Mildenhall B., Verbin D., Srinivasan P.~P., Hedman P.}:
\newblock Mip-nerf 360: Unbounded anti-aliased neural radiance fields.
\newblock In \emph{Proceedings of the IEEE/CVF Conference on Computer Vision and Pattern Recognition} (2022), pp.~5470--5479.

\bibitem[BZY{\etalchar{*}}23]{n_sine}
\textsc{Bao C., Zhang Y., Yang B., Fan T., Yang Z., Bao H., Zhang G., Cui Z.}:
\newblock Sine: Semantic-driven image-based nerf editing with prior-guided editing field.
\newblock In \emph{Proceedings of the IEEE/CVF Conference on Computer Vision and Pattern Recognition} (2023), pp.~20919--20929.

\bibitem[CBCP20]{c_replay1}
\textsc{Caccia L., Belilovsky E., Caccia M., Pineau J.}:
\newblock Online learned continual compression with adaptive quantization modules.
\newblock In \emph{International conference on machine learning} (2020), PMLR, pp.~1240--1250.

\bibitem[CCW{\etalchar{*}}23]{slam-r-1}
\textsc{Chen Y., Chen X., Wang X., Zhang Q., Guo Y., Shan Y., Wang F.}:
\newblock Local-to-global registration for bundle-adjusting neural radiance fields.
\newblock In \emph{Proceedings of the IEEE/CVF Conference on Computer Vision and Pattern Recognition} (2023), pp.~8264--8273.

\bibitem[CFHT23]{n_mobilenerf}
\textsc{Chen Z., Funkhouser T., Hedman P., Tagliasacchi A.}:
\newblock Mobilenerf: Exploiting the polygon rasterization pipeline for efficient neural field rendering on mobile architectures.
\newblock In \emph{Proceedings of the IEEE/CVF Conference on Computer Vision and Pattern Recognition} (2023), pp.~16569--16578.

\bibitem[CLBL22]{n_meil_nerf}
\textsc{Chung J., Lee K., Baik S., Lee K.~M.}:
\newblock Meil-nerf: Memory-efficient incremental learning of neural radiance fields, 2022.
\newblock \href {http://arxiv.org/abs/2212.08328} {\path{arXiv:2212.08328}}.

\bibitem[CXG{\etalchar{*}}22]{n_tensorf}
\textsc{Chen A., Xu Z., Geiger A., Yu J., Su H.}:
\newblock Tensorf: Tensorial radiance fields.
\newblock In \emph{European Conference on Computer Vision} (2022), Springer, pp.~333--350.

\bibitem[CXZ{\etalchar{*}}21]{n_mvsnerf}
\textsc{Chen A., Xu Z., Zhao F., Zhang X., Xiang F., Yu J., Su H.}:
\newblock Mvsnerf: Fast generalizable radiance field reconstruction from multi-view stereo.
\newblock In \emph{Proceedings of the IEEE/CVF International Conference on Computer Vision} (2021), pp.~14124--14133.

\bibitem[CYM{\etalchar{*}}23]{n_gm_nerf}
\textsc{Chen J., Yi W., Ma L., Jia X., Lu H.}:
\newblock Gm-nerf: Learning generalizable model-based neural radiance fields from multi-view images.
\newblock In \emph{Proceedings of the IEEE/CVF Conference on Computer Vision and Pattern Recognition} (2023), pp.~20648--20658.

\bibitem[DSQ{\etalchar{*}}24]{slam-r-2}
\textsc{Deng T., Shen G., Qin T., Wang J., Zhao W., Wang J., Wang D., Chen W.}:
\newblock Plgslam: Progressive neural scene represenation with local to global bundle adjustment.
\newblock In \emph{Proceedings of the IEEE/CVF Conference on Computer Vision and Pattern Recognition} (2024), pp.~19657--19666.

\bibitem[EMC{\etalchar{*}}20]{c_arch2}
\textsc{Ebrahimi S., Meier F., Calandra R., Darrell T., Rohrbach M.}:
\newblock Adversarial continual learning.
\newblock In \emph{Computer Vision--ECCV 2020: 16th European Conference, Glasgow, UK, August 23--28, 2020, Proceedings, Part XI 16} (2020), Springer, pp.~386--402.

\bibitem[FAL17]{meta-maml}
\textsc{Finn C., Abbeel P., Levine S.}:
\newblock Model-agnostic meta-learning for fast adaptation of deep networks.
\newblock In \emph{International conference on machine learning} (2017), PMLR, pp.~1126--1135.

\bibitem[FR22]{meta-metaappearance}
\textsc{Fischer M., Ritschel T.}:
\newblock Metappearance: Meta-learning for visual appearance reproduction.
\newblock \emph{ACM Transactions on Graphics (TOG) 41}, 6 (2022), 1--13.

\bibitem[FXW{\etalchar{*}}23]{n_pvd}
\textsc{Fang S., Xu W., Wang H., Yang Y., Wang Y., Zhou S.}:
\newblock One is all: Bridging the gap between neural radiance fields architectures with progressive volume distillation.
\newblock In \emph{Proceedings of the AAAI Conference on Artificial Intelligence} (2023), vol.~37, pp.~597--605.

\bibitem[GCML23]{n_quantizing_nerf}
\textsc{Gordon C., Chng S.-F., MacDonald L., Lucey S.}:
\newblock On quantizing implicit neural representations.
\newblock In \emph{Proceedings of the IEEE/CVF Winter Conference on Applications of Computer Vision} (2023), pp.~341--350.

\bibitem[GPAM{\etalchar{*}}14]{gan}
\textsc{Goodfellow I., Pouget-Abadie J., Mirza M., Xu B., Warde-Farley D., Ozair S., Courville A., Bengio Y.}:
\newblock Generative adversarial nets.
\newblock \emph{Advances in neural information processing systems 27} (2014).

\bibitem[HDL16]{hypernetworks}
\textsc{Ha D., Dai A.~M., Le Q.~V.}:
\newblock Hypernetworks.
\newblock \emph{CoRR abs/1609.09106} (2016).
\newblock URL: \url{http://arxiv.org/abs/1609.09106}, \href {http://arxiv.org/abs/1609.09106} {\path{arXiv:1609.09106}}.

\bibitem[HLX{\etalchar{*}}23]{cp-nerf}
\textsc{He H., Liang Y., Xiao S., Chen J., Chen Y.}:
\newblock Cp-nerf: Conditionally parameterized neural radiance fields for cross-scene novel view synthesis.
\newblock In \emph{Computer Graphics Forum} (2023), vol.~42, Wiley Online Library, p.~e14940.

\bibitem[HSW{\etalchar{*}}21]{lora}
\textsc{Hu E.~J., Shen Y., Wallis P., Allen-Zhu Z., Li Y., Wang S., Wang L., Chen W.}:
\newblock Lora: Low-rank adaptation of large language models.
\newblock \emph{arXiv preprint arXiv:2106.09685} (2021).

\bibitem[HZRS16]{resnet}
\textsc{He K., Zhang X., Ren S., Sun J.}:
\newblock Deep residual learning for image recognition.
\newblock In \emph{Proceedings of the IEEE conference on computer vision and pattern recognition} (2016), pp.~770--778.

\bibitem[KB14]{exp_adam}
\textsc{Kingma D.~P., Ba J.}:
\newblock Adam: A method for stochastic optimization.
\newblock \emph{arXiv preprint arXiv:1412.6980} (2014).

\bibitem[KGC17]{mtl_loss}
\textsc{Kendall A., Gal Y., Cipolla R.}:
\newblock Multi-task learning using uncertainty to weigh losses for scene geometry and semantics.
\newblock \emph{CoRR abs/1705.07115} (2017).
\newblock URL: \url{http://arxiv.org/abs/1705.07115}, \href {http://arxiv.org/abs/1705.07115} {\path{arXiv:1705.07115}}.

\bibitem[KKZS23]{n_hypernerfgan}
\textsc{Kania A., Kasymov A., Zięba M., Spurek P.}:
\newblock Hypernerfgan: Hypernetwork approach to 3d nerf gan, 2023.
\newblock \href {http://arxiv.org/abs/2301.11631} {\path{arXiv:2301.11631}}.

\bibitem[KPR{\etalchar{*}}17]{c_reg1}
\textsc{Kirkpatrick J., Pascanu R., Rabinowitz N., Veness J., Desjardins G., Rusu A.~A., Milan K., Quan J., Ramalho T., Grabska-Barwinska A., et~al.}:
\newblock Overcoming catastrophic forgetting in neural networks.
\newblock \emph{Proceedings of the national academy of sciences 114}, 13 (2017), 3521--3526.

\bibitem[KPZK17]{exp_tt}
\textsc{Knapitsch A., Park J., Zhou Q.-Y., Koltun V.}:
\newblock Tanks and temples: Benchmarking large-scale scene reconstruction.
\newblock \emph{ACM Transactions on Graphics 36}, 4 (2017).

\bibitem[LGL{\etalchar{*}}20]{n_nsvf}
\textsc{Liu L., Gu J., Lin K.~Z., Chua T.-S., Theobalt C.}:
\newblock Neural sparse voxel fields.
\newblock \emph{NeurIPS} (2020).

\bibitem[LMW21]{n_autoint}
\textsc{Lindell D.~B., Martel J.~N., Wetzstein G.}:
\newblock Autoint: Automatic integration for fast neural volume rendering.
\newblock In \emph{Proceedings of the IEEE/CVF Conference on Computer Vision and Pattern Recognition} (2021), pp.~14556--14565.

\bibitem[LSW{\etalchar{*}}23]{n_vqnerf}
\textsc{Li L., Shen Z., Wang Z., Shen L., Bo L.}:
\newblock Compressing volumetric radiance fields to 1 mb.
\newblock In \emph{Proceedings of the IEEE/CVF Conference on Computer Vision and Pattern Recognition} (2023), pp.~4222--4231.

\bibitem[LWM{\etalchar{*}}20]{c_replay3}
\textsc{Liu X., Wu C., Menta M., Herranz L., Raducanu B., Bagdanov A.~D., Jui S., de~Weijer J.~v.}:
\newblock Generative feature replay for class-incremental learning.
\newblock In \emph{Proceedings of the IEEE/CVF Conference on Computer Vision and Pattern Recognition Workshops} (2020), pp.~226--227.

\bibitem[LZQ{\etalchar{*}}22]{c_arch3}
\textsc{Lin H., Zhang Y., Qiu Z., Niu S., Gan C., Liu Y., Tan M.}:
\newblock Prototype-guided continual adaptation for class-incremental unsupervised domain adaptation.
\newblock In \emph{European Conference on Computer Vision} (2022), Springer, pp.~351--368.

\bibitem[MBRS{\etalchar{*}}21]{nerf_w}
\textsc{Martin-Brualla R., Radwan N., Sajjadi M.~S., Barron J.~T., Dosovitskiy A., Duckworth D.}:
\newblock Nerf in the wild: Neural radiance fields for unconstrained photo collections.
\newblock In \emph{Proceedings of the IEEE/CVF Conference on Computer Vision and Pattern Recognition} (2021), pp.~7210--7219.

\bibitem[MDL18]{c_arch1}
\textsc{Mallya A., Davis D., Lazebnik S.}:
\newblock Piggyback: Adapting a single network to multiple tasks by learning to mask weights.
\newblock In \emph{Proceedings of the European conference on computer vision (ECCV)} (2018), pp.~67--82.

\bibitem[MESK22]{n_ingp}
\textsc{M{\"u}ller T., Evans A., Schied C., Keller A.}:
\newblock Instant neural graphics primitives with a multiresolution hash encoding.
\newblock \emph{ACM Transactions on Graphics (ToG) 41}, 4 (2022), 1--15.

\bibitem[ML18]{packnet}
\textsc{Mallya A., Lazebnik S.}:
\newblock Packnet: Adding multiple tasks to a single network by iterative pruning.
\newblock In \emph{Proceedings of the IEEE conference on Computer Vision and Pattern Recognition} (2018), pp.~7765--7773.

\bibitem[MLP{\etalchar{*}}21]{exp_relu}
\textsc{Mishra A., Latorre J.~A., Pool J., Stosic D., Stosic D., Venkatesh G., Yu C., Micikevicius P.}:
\newblock Accelerating sparse deep neural networks, 2021.
\newblock \href {http://arxiv.org/abs/2104.08378} {\path{arXiv:2104.08378}}.

\bibitem[MSOC{\etalchar{*}}19]{exp_llff}
\textsc{Mildenhall B., Srinivasan P.~P., Ortiz-Cayon R., Kalantari N.~K., Ramamoorthi R., Ng R., Kar A.}:
\newblock Local light field fusion: Practical view synthesis with prescriptive sampling guidelines.
\newblock \emph{ACM Transactions on Graphics (TOG) 38}, 4 (2019), 1--14.

\bibitem[MST{\etalchar{*}}21]{n_vanilla_nerf}
\textsc{Mildenhall B., Srinivasan P.~P., Tancik M., Barron J.~T., Ramamoorthi R., Ng R.}:
\newblock Nerf: Representing scenes as neural radiance fields for view synthesis.
\newblock \emph{Communications of the ACM 65}, 1 (2021), 99--106.

\bibitem[PDBW23]{icl-nerf}
\textsc{Po R., Dong Z., Bergman A.~W., Wetzstein G.}:
\newblock Instant continual learning of neural radiance fields.
\newblock In \emph{Proceedings of the IEEE/CVF International Conference on Computer Vision} (2023), pp.~3334--3344.

\bibitem[PYS{\etalchar{*}}23]{n_mlpmaps}
\textsc{Peng S., Yan Y., Shuai Q., Bao H., Zhou X.}:
\newblock Representing volumetric videos as dynamic mlp maps, 2023.
\newblock \href {http://arxiv.org/abs/2304.06717} {\path{arXiv:2304.06717}}.

\bibitem[QRX{\etalchar{*}}21]{c_survey1}
\textsc{Qu H., Rahmani H., Xu L., Williams B., Liu J.}:
\newblock Recent advances of continual learning in computer vision: An overview.
\newblock \emph{arXiv preprint arXiv:2109.11369} (2021).

\bibitem[RKSL17]{c_replay2}
\textsc{Rebuffi S.-A., Kolesnikov A., Sperl G., Lampert C.~H.}:
\newblock icarl: Incremental classifier and representation learning.
\newblock In \emph{Proceedings of the IEEE conference on Computer Vision and Pattern Recognition} (2017), pp.~2001--2010.

\bibitem[RRD{\etalchar{*}}16]{c_arch4}
\textsc{Rusu A.~A., Rabinowitz N.~C., Desjardins G., Soyer H., Kirkpatrick J., Kavukcuoglu K., Pascanu R., Hadsell R.}:
\newblock Progressive neural networks.
\newblock \emph{arXiv preprint arXiv:1606.04671} (2016).

\bibitem[SCT{\etalchar{*}}20]{meta-metasdf}
\textsc{Sitzmann V., Chan E., Tucker R., Snavely N., Wetzstein G.}:
\newblock Metasdf: Meta-learning signed distance functions.
\newblock \emph{Advances in Neural Information Processing Systems 33} (2020), 10136--10147.

\bibitem[SLOD21]{imap}
\textsc{Sucar E., Liu S., Ortiz J., Davison A.~J.}:
\newblock imap: Implicit mapping and positioning in real-time.
\newblock In \emph{Proceedings of the IEEE/CVF International Conference on Computer Vision} (2021), pp.~6229--6238.

\bibitem[SP23]{n_binarynerf}
\textsc{Shin S., Park J.}:
\newblock Binary radiance fields.
\newblock \emph{arXiv preprint arXiv:2306.07581} (2023).

\bibitem[SSA{\etalchar{*}}23]{n_hypnerf}
\textsc{Sen B., Singh G., Agarwal A., Agaram R., Krishna K.~M., Sridhar S.}:
\newblock Hyp-nerf: Learning improved nerf priors using a hypernetwork, 2023.
\newblock \href {http://arxiv.org/abs/2306.06093} {\path{arXiv:2306.06093}}.

\bibitem[SSC22]{n_dvgo}
\textsc{Sun C., Sun M., Chen H.-T.}:
\newblock Direct voxel grid optimization: Super-fast convergence for radiance fields reconstruction.
\newblock In \emph{Proceedings of the IEEE/CVF Conference on Computer Vision and Pattern Recognition} (2022), pp.~5459--5469.

\bibitem[TCWZ22]{n_ccnerf}
\textsc{Tang J., Chen X., Wang J., Zeng G.}:
\newblock Compressible-composable nerf via rank-residual decomposition.
\newblock \emph{Advances in Neural Information Processing Systems 35} (2022), 14798--14809.

\bibitem[TMW{\etalchar{*}}21]{meta-lll}
\textsc{Tancik M., Mildenhall B., Wang T., Schmidt D., Srinivasan P.~P., Barron J.~T., Ng R.}:
\newblock Learned initializations for optimizing coordinate-based neural representations.
\newblock In \emph{Proceedings of the IEEE/CVF Conference on Computer Vision and Pattern Recognition} (2021), pp.~2846--2855.

\bibitem[VSP{\etalchar{*}}17]{transformer}
\textsc{Vaswani A., Shazeer N., Parmar N., Uszkoreit J., Jones L., Gomez A.~N., Kaiser {\L}., Polosukhin I.}:
\newblock Attention is all you need.
\newblock \emph{Advances in neural information processing systems 30} (2017).

\bibitem[WRB{\etalchar{*}}23]{n_duplexnerf}
\textsc{Wan Z., Richardt C., Bo{\v{z}}i{\v{c}} A., Li C., Rengarajan V., Nam S., Xiang X., Li T., Zhu B., Ranjan R., et~al.}:
\newblock Learning neural duplex radiance fields for real-time view synthesis.
\newblock In \emph{Proceedings of the IEEE/CVF Conference on Computer Vision and Pattern Recognition} (2023), pp.~8307--8316.

\bibitem[WWG{\etalchar{*}}21]{n_ibrnet}
\textsc{Wang Q., Wang Z., Genova K., Srinivasan P.~P., Zhou H., Barron J.~T., Martin-Brualla R., Snavely N., Funkhouser T.}:
\newblock Ibrnet: Learning multi-view image-based rendering.
\newblock In \emph{Proceedings of the IEEE/CVF Conference on Computer Vision and Pattern Recognition} (2021), pp.~4690--4699.

\bibitem[WWQQ23]{n_ripnerf}
\textsc{Wang Y., Wang J., Qu Y., Qi Y.}:
\newblock Rip-nerf: Learning rotation-invariant point-based neural radiance field for fine-grained editing and compositing.
\newblock In \emph{Proceedings of the 2023 ACM International Conference on Multimedia Retrieval} (2023), pp.~125--134.

\bibitem[WZSZ23]{c_survey2}
\textsc{Wang L., Zhang X., Su H., Zhu J.}:
\newblock A comprehensive survey of continual learning: Theory, method and application.
\newblock \emph{arXiv preprint arXiv:2302.00487} (2023).

\bibitem[XXP{\etalchar{*}}22]{bungeenerf}
\textsc{Xiangli Y., Xu L., Pan X., Zhao N., Rao A., Theobalt C., Dai B., Lin D.}:
\newblock Bungeenerf: Progressive neural radiance field for extreme multi-scale scene rendering.
\newblock In \emph{European conference on computer vision} (2022), Springer, pp.~106--122.

\bibitem[YHL{\etalchar{*}}23]{n_contranerf}
\textsc{Yang H., Hong L., Li A., Hu T., Li Z., Lee G.~H., Wang L.}:
\newblock Contranerf: Generalizable neural radiance fields for synthetic-to-real novel view synthesis via contrastive learning.
\newblock In \emph{Proceedings of the IEEE/CVF Conference on Computer Vision and Pattern Recognition} (2023), pp.~16508--16517.

\bibitem[YZX{\etalchar{*}}21]{object_nerf}
\textsc{Yang B., Zhang Y., Xu Y., Li Y., Zhou H., Bao H., Zhang G., Cui Z.}:
\newblock Learning object-compositional neural radiance field for editable scene rendering.
\newblock In \emph{Proceedings of the IEEE/CVF International Conference on Computer Vision} (2021), pp.~13779--13788.

\bibitem[ZC23]{clnerf}
\textsc{Zhipeng~Cai M.~M.}:
\newblock Clnerf: Continual learning meets nerf.
\newblock In \emph{ICCV} (2023).

\bibitem[ZLCX23]{il-nerf}
\textsc{Zhang L., Li M., Chen C., Xu J.}:
\newblock Il-nerf: Incremental learning for neural radiance fields with camera pose alignment.
\newblock \emph{arXiv preprint arXiv:2312.05748} (2023).

\bibitem[ZSK{\etalchar{*}}19]{meta-cavia}
\textsc{Zintgraf L., Shiarli K., Kurin V., Hofmann K., Whiteson S.}:
\newblock Fast context adaptation via meta-learning.
\newblock In \emph{International Conference on Machine Learning} (2019), PMLR, pp.~7693--7702.

\end{thebibliography}



\newpage

\begin{LARGE}  
\textbf{Supplementary material\\}  
\end{LARGE}
\appendix
\setcounter{figure}{0}
\setcounter{table}{0}
\renewcommand\thesection{\Alph{section}}
\renewcommand\thetable{\Alph{table}}
\renewcommand\thefigure{\Alph{figure}}
In this supplementary material, we describe more details of our method, including model architecture in Sec. A, more implementation details in Sec.B, and more discussions in Sec.C. Besides, we also provide more experiment results in Sec.D.
\section{Model Architecture}
\begin{figure}[htb]
\centering
\includegraphics[width=1.0\columnwidth]{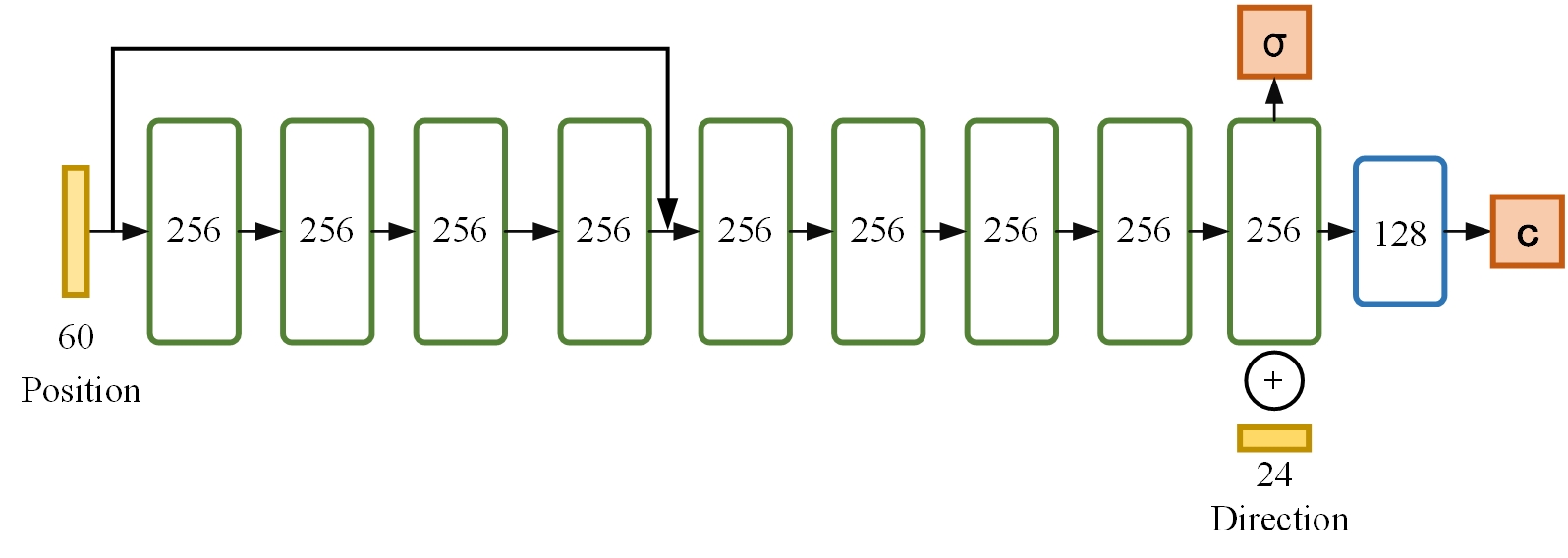}
\caption{The model architecture of SCARF.}
\label{sup_arch_nerf}
\end{figure}

The detailed model architecture is shown in Fig. \ref{sup_arch_nerf}. Input vectors are shown in yellow, output vectors are shown in orange, generated hidden layers are shown in green, and global cross-scene decoder layers are shown in blue. Both the generated layers and the decoder layers are fully-connected layers with a ReLU \cite{exp_relu} activation. The number inside each block signifies the vector's dimension, and "+" denotes the concatenation operation. To begin with, the 10 degree positional encoding of the input position $x$ is passed through 8 generated fully-connected layers, each with 256 channels. We follow the vanilla NeRF architecture to include a skip connection at the fifth layer. An additional layer outputs the density $\sigma$ and a 256-dimensional feature vector. This feature is concatenated with the 4 degree positional encoding of the input viewing direction processed by a cross-scene decoder with 128 channels and 3 channels to output the final RGB radiance. For different 3D scenes, different parameters of the encoder of are generated, thus achieving a multi-scene representation.

\begin{figure}[htb]
\centering
\includegraphics[width=1.0\columnwidth]{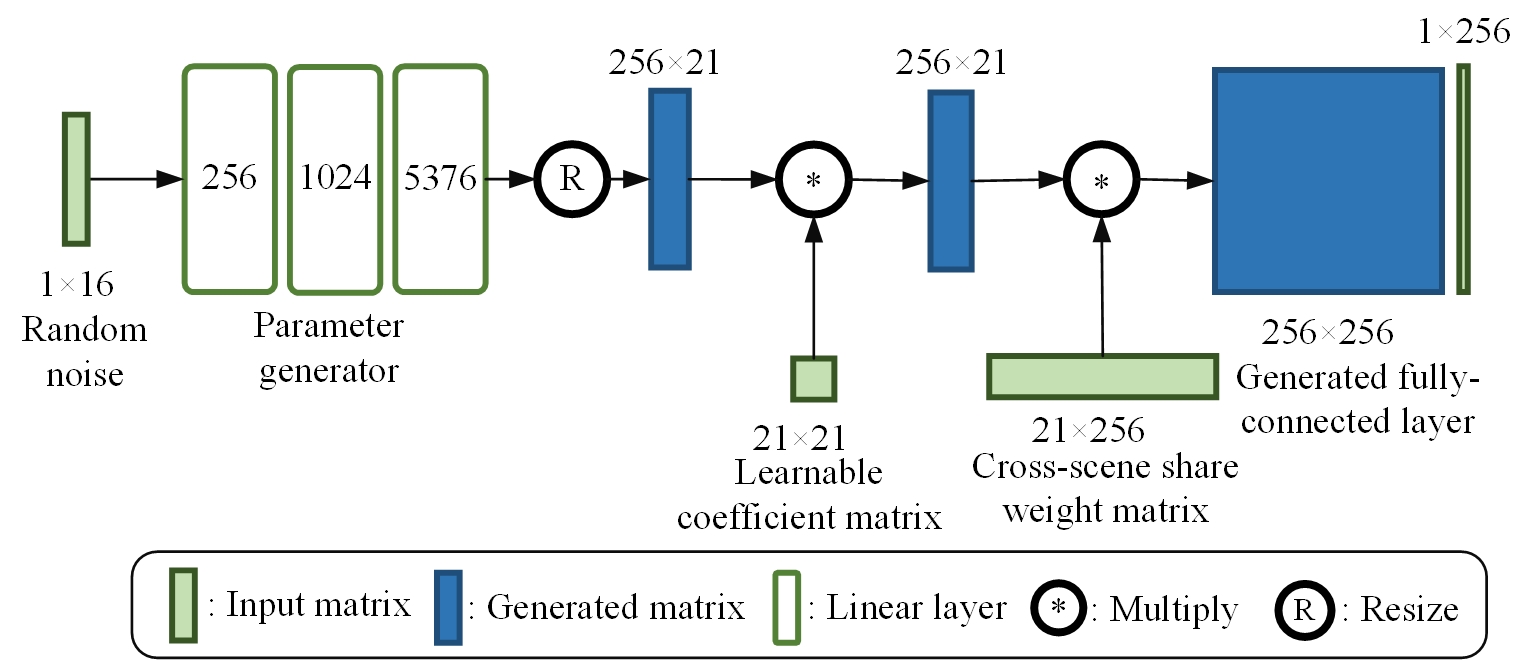}
\caption{The model architecture of a fully-connected layer parameter generation process in SCARF, in which the input and output dim are both 256.}
\label{sup_arch}
\end{figure}

Fig. \ref{sup_arch} demonstrates the model architecture of a fully-connected layer parameter generation process in SCARF, in which the input and output dim are both 256. Input random noise or learnable matrices are shown in green, and generated matrices are shown in blue. For a specific scene, random noise is input to the parameter generator to output a latent feature. After resizing the feature to $ 256 \times 21$, A learnable coefficient matrix is multiplied by the feature. Finally, by multiplying the cross-scene share weight matrix in which the shape is $ 21 \times 256$, the parameter of fully-connected layer is generated.

\section{More Implementation Details}
\subsection{More Training Details}
As introduced in our main paper, we adopt the USD strategy to distill the knowledge of previous 3D scenes. We adopt the Adam optimizer with the initial learning rates of $5e-4$ for the learnable matrices (contains SSWM, CSWM, coefficient matrix), and decays exponentially to $5e-5$ over the optimization. The initial learning rate is set as $1e-4$ for the parameters generator and decays exponentially to $5e-5$ over the optimization. We optimize the upcoming new scene with a batch size of 4096 pixel rays. We distill the knowledge of previously learned scenes with a batch size of 1024 pixel rays and 8192 random sampled points. Moreover, we found that warm-up training for new scenes gives better results. So, we train the model only with the loss of $\mathcal{L}_{C(r)}^t$ for 2K steps, then joint training all scenes with all loss proposed in the main paper. Besides, to train on the TanksAndTemples \cite{exp_tt} dataset that contains unbounded background for the ground truth RGB images, we follow the NSVF \cite{n_nsvf} to mask the foreground.
\subsection{Details of USD density occupancy grid extracting.}
The resolution of the occupancy grid is set to $50 \times 50 \times 50$. During the surface distillation, we divide each grid cell into a subgrid of $5 \times 5 \times 5$ and query the density of each subgrid point. A volume cell is considered as the surface for distillation if there exists a subgrid point inside with density above the threshold $\tau$ ($\tau=3$ in all our experiments).
\subsection{Details of other methods}
\subsubsection{Details of EWC with NeRF}
Since the regularization loss is generally small, we multiply the loss by a large weight, as in the original EWC paper \cite{c_reg1}. For every dataset, we use $10^8$ for the weight, making the regularization loss to be approximately 10\% of the photo-metric loss in vanilla NeRF \cite{n_vanilla_nerf}.
\subsubsection{Details of PackNet with NeRF}
MEIL-NeRF \cite{n_meil_nerf} also compared with PackNet \cite{packnet} that reconstruction in a single scene with multiple sequences. We refer the settings in MEIL-NeRF, and apply the PackNet to multiple scenes continual learning scenario. We set the pruning rate to 0.5 to follow the settings of the original paper while clearly showing the characteristics of PackNet in NeRF.
\subsubsection{Details of MEIL-NeRF and CLNeRF}

MEIL-NeRF \cite{n_meil_nerf} is designed for continual learning a single scenes with multiple sequences. CLNeRF \cite{clnerf} is designed for continual learning a single scene with a sequence of multiple scans with appearance and geometry changes, over an extended period of time. Although they both try to introduce continual learning into NeRF. They can't handle the multiple scene continual learning scenario. For fare comparison, we improve their network architecture with a 32 diminsion scene-specific learnable latent code and optimize them during the multiple scene continual learning. The Generative Latent Optimization (GLO) is widely used in decompose multiple NeRF with a single network \cite{nerf_w,object_nerf}.

\section{More Discussions}
\subsection{Number of parameters}
We briefly discuss the number of parameters in our model. With the same NeRF architecture, vanilla NeRF uses fully-connected layer to represent the scene. The total number of parameters used for vanilla NeRF is approximate $(c_{in} \times (c_{out}+1) \times L) \times N + P_{d} \times N$, while $c_{in}$ and $c_{out}$ are the input and output dim for each fully-connected layer of the encoder, $L$ is the number of encoder's layer, $P_{d}$ is the number of the decoder's parameters, and $N$ is the number of the learned scenes. Our SCARF factorizes the NeRF model into a sets of SSWM and a global CSWM. and the sets of SSWM is generated from a global parameter generator with the scene-specific random noise. The total number of parameters used for SCARF is approximately $N \times( Z + P_c ) + L \times (P_G + P_{CSWM}) +P_{d} \approx N \times( Z + K\times K ) + L \times (Z\times K \times c_{in} + K \times c_{out}) +P_{d}$. $Z$ is the parameters number of random noise, $P_c$ is the parameter number of coefficient matrix, $P_G$ is the parameter number of parameter generator $G(.)$, $P_{CSWM}$ is the parameter number of CSWM. The number of parameters of the network in vanilla NeRF grows linearly with the number of scenes $N$. In contrast, in SCARF, parameters depend most on the number of layers $L$ of model. As we discuss in the main paper, additional scene training requires only very few parameters. Theoretically, our SCARF model can achieve $16.4\%$ and $0.03\%$ compression rates for 8 scenes and 50 scenes respectively compared with vanilla NeRF.

\subsection{About the training/rendering time, and about the generalizability of the proposed SCARF.}
In the rendering stage, we pre-generate the weights of MLPs for NeRF with cross-scene weight matrix and parameters generator. So, we do not need any additional computation at the rendering time, and \textbf{the rendering speed is equivalent to the vanilla NeRF}. In the training phase, although the parameters for SCARF (the proposed method) and vanilla NeRF are comparable, SCARF needs to perform the USD (Uncertain Surface Distillation) strategy to distill the learned scenes knowledge from previous learned model, which requires some additional time. \textbf{SCARF is about 1.2 $\times$ slower than vanilla NeRF in training phase}. 

\textbf{Note that we do not use more advanced baselines} such as fast rendering NeRF, fast convergence NeRF, editable NeRF, etc. Because SCARF explores the possibility of memory-efficient representation and continuous learning across multiple scenes. \textbf{It is a plug-and-play approach. When the parameters of NeRF are generated by SCARF, the other pipeline (sampling, rendering, etc.) is same as NeRF, and can be combined with other NeRF extension work seamlessly.} Based on vanilla NeRF to construct our framework provides a fairer reflection of the effectiveness about the memory-efficient and continual learning for multiple scenes, and more generalizability for future work. We consider that SCARF is orthogonal to other single-scene NeRF extension works and can be combined in the future.

\subsection{Relation to NeRF with HyperNetworks methods}
Some very recent researches also combined NeRF with HyperNetworks \cite{hypernetworks}. HyP-NeRF \cite{n_hypnerf} learns a category-level NeRF to achieve single-view input and multiple view rendering. HyperNeRFGAN \cite{n_hypernerfgan} proposes a generative model and uses the HypernetWorks paradigm to produce 3D objects represented by NeRF. MLP-Maps \cite{n_mlpmaps} represents the volume video as a set of shallow MLP networks whose parameters are stored in 2D grids. These works likewise demonstrate that generating the network parameters of the NeRF through another hypernetwork helps improve the generalization of the NeRF. However, our design solutions of the HyperNetworks, and the goals we pursue are different from theirs.

\subsection{Relation to BungeeNeRF}
BungeeNeRF \cite{bungeenerf} focuses on multi-scale NeRF representation of a single scene, preserving high-quality details across scales from satellite to ground level in an outdoor scene. Thus, BungeeNeRF's challenge lies in representing the scene hierarchically with different NeRF models, where different NeRFs represent various frequency signals from low to high for a single scene. In contrast, our proposed work explores compact NeRF representations across different scenes. We demonstrate that the NeRF model can be learned with low-rank weight matrices for the MLP, showing that the parameters of NeRF models learned from different scenes with significant domain gaps can be efficiently shared and learned continuously.

\section{More Experiment Results}
\begin{figure}[htbp]
\centering
\includegraphics[width=1.0\columnwidth]{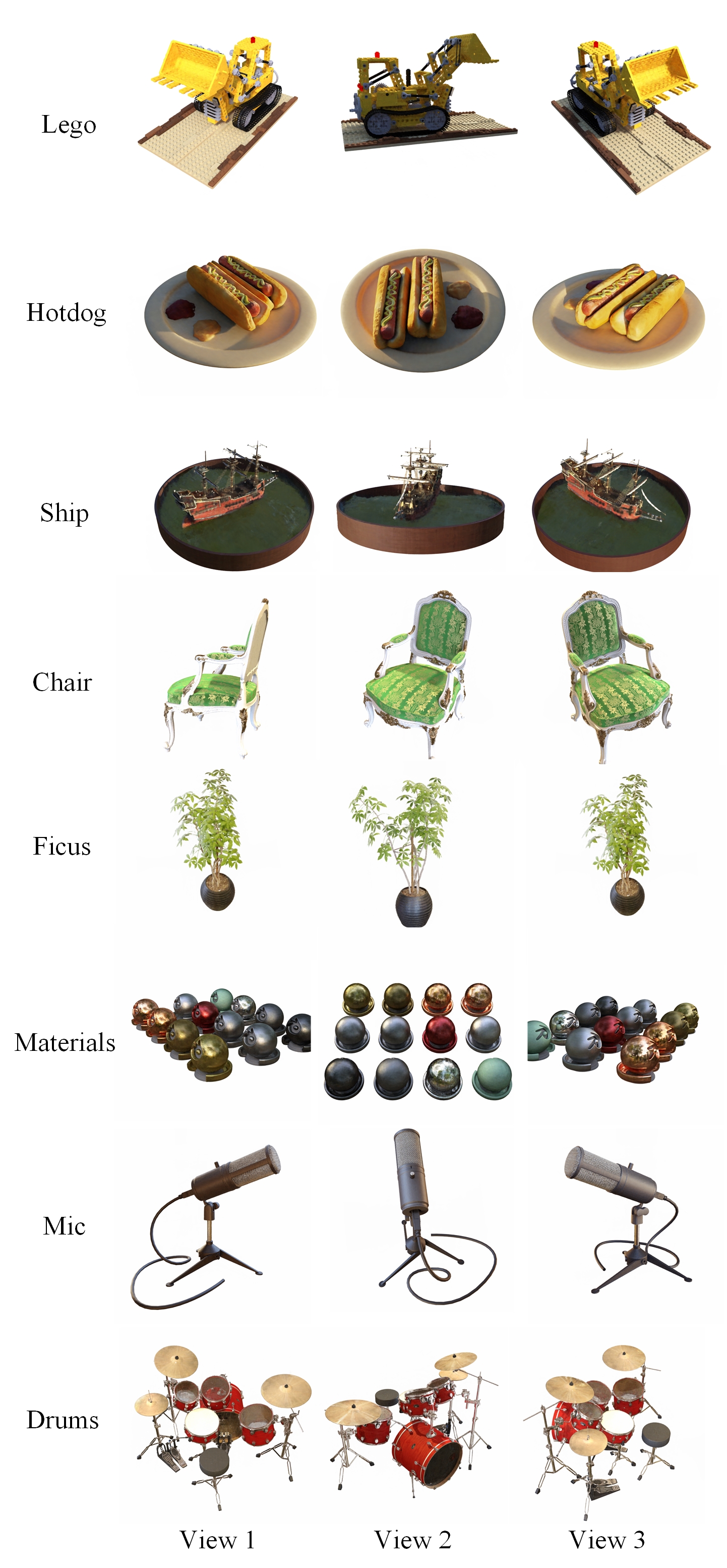}
\caption{Qualitative results of continual learning of eight scenes on NeRF-Synthetic dataset.}
\label{sup_vis_nerf}
\end{figure}

Tab.\ref{table_sup_tt}, Tab.\ref{table_sup_llff}, and Tab.\ref{table_sup_nerf} give concrete quantitative results of continual learning on the scenes from  TanksAndTemples \cite{exp_tt}, LLFF \cite{exp_llff}, and NeRF-Synthetic \cite{n_vanilla_nerf} datasets, respectively. Fig.\ref{sup_vis_nerf} gives concrete qualitative results of continual learning on the scenes from NeRF-Synthetic \cite{n_vanilla_nerf} datasets.

\begin{table}[htbp]
\resizebox{1.0\columnwidth}{!}{
\begin{tabular}{ccccccc}
\hline

    Metric &   Barn      & Caterpillar      & Family      & Ignatius &Truck   &Avg.                     \\ \hline
  PSNR&    25.51&25.03&32.47&25.41&25.49&26.78\\
SSIM& 0.795&0.886&0.942&0.914&0.922&0.892\\
\hline
\end{tabular}}
\caption{Quantitative results of continual learning on TanksAndTemples dataset. We calculate the PSNR and SSIM after continual learning five scenes in sequence.}
\label{table_sup_tt}
\end{table}

\begin{table}[htbp]
\resizebox{1.0\columnwidth}{!}{
\begin{tabular}{c|cccccccc|c}
\hline
 Metric &    Fern      & Flower      & Fortress      & Horns &Leaves &Orchids &Room &T-Rex  &Avg. \\  \hline
   PSNR&    25.00 &27.21&31.12&27.38&20.89&20.31 &32.67&26.97& 26.44\\
      SSIM& 0.785 &0.824&0.876&0.823&0.689&0.638&0.945&0.882&0.808\\
\hline
\end{tabular}}
\caption{Quantitative results of continual learning on LLFF dataset. We calculate the PSNR and SSIM after continual learning eight scenes in sequence.}
\label{table_sup_llff}
\end{table}
\begin{table}[htbp]
\resizebox{1.0\columnwidth}{!}{
\begin{tabular}{cccccccccc}
\hline
 Metric &    Hotdog      & Lego      & Ship      & Chair &Ficus &Materials &Mic &Drums  &Avg.                     \\ \hline
   PSNR&    36.01 &32.39&28.62&33.15& 30.19&29.29 &32.89&24.94& 30.94\\
      SSIM& 0.969 &0.953&0.852&0.962&0.960&0.941&0.970&0.956&0.945\\
\hline
\end{tabular}}
\caption{Quantitative results of continual learning on NeRF-Synthetic dataset. We calculate the PSNR and SSIM after continual learning eight scenes in sequence.}
\label{table_sup_nerf}
\end{table}

\end{document}